\documentclass[journal,twoside,web]{ieeecolor}
\usepackage{tmi}
\usepackage{cite}
\usepackage{amsmath,amssymb,amsfonts}
\usepackage{algorithmic}
\usepackage{graphicx}
\usepackage{textcomp}
\usepackage{colortbl}
\usepackage{bbm}
\usepackage{amsmath}
\usepackage{mathtools}
\usepackage{makeidx}  % allows for indexgeneration
\usepackage{subfig}
\usepackage{graphicx}
\usepackage{amssymb,amsmath,bm}
\usepackage{booktabs}
\usepackage[misc]{ifsym}
\usepackage{multirow}
\usepackage{bbding}
\usepackage{array}
\usepackage{float}
\usepackage{algorithm}
\usepackage{algorithmic}
\usepackage{lettrine}
\usepackage{cite}
\usepackage{xcolor}
\colorlet{blue}{black}
\usepackage{mathrsfs}
\usepackage[switch]{lineno}
\usepackage{xspace}
\newcommand{\etal}{{\emph{et al.}}}
\newcommand{\eg}{{\emph{e.g.}}}
\newcommand{\ie}{{\emph{i.e.}}}

\newcolumntype{C}[1]{>{\centering\let\newline\\\arraybackslash}m{#1}}
\renewcommand\arraystretch{1.5}
\begin{document}%\linenumbers
\title{Domain Adaptation Meets Zero-Shot Learning: An Annotation-Efficient Approach to Multi-Modality Medical Image Segmentation}
\author{Cheng Bian, Chenglang Yuan, Kai Ma, Shuang Yu, Dong Wei and Yefeng Zheng,~\IEEEmembership{Fellow,~IEEE}% <-this % stops a space
    \thanks{This work was supported by the National Key R\&D Program of China (2020AAA0109500/2020AAA0109501).
    \textit{Cheng Bian and Chenglang Yuan contribute equally to this work.}
    \textit{Corresponding author: Dong Wei (donwei@tencent.com).}}
    \thanks{Cheng Bian, Chenglang Yuan, Kai Ma, and Shuang Yu are with the Tencent Jarvis Lab, Shenzhen 518057, China.}
    \thanks{Dong Wei and Yefeng Zheng are with the Tencent Healthcare (Shenzhen) Co., LTD.}}

\maketitle

\begin{abstract}
Due to the lack of properly annotated medical data, exploring the generalization capability of the deep model is becoming a public concern.
Zero-shot learning (ZSL) has emerged in recent years to equip the deep model with the ability to recognize unseen classes. However, existing studies mainly focus on natural images, which utilize linguistic models to extract auxiliary information for ZSL. It is impractical to apply the natural image ZSL solutions directly to medical images, since the medical terminology is very domain-specific, and it is not easy to acquire linguistic models for the medical terminology. In this work, we propose a new paradigm of ZSL specifically for medical images utilizing cross-modality information.
We make three main contributions with the proposed paradigm. First, we extract the prior knowledge about the segmentation targets, called relation prototypes, from the prior model and then propose a cross-modality adaptation module to inherit the prototypes to the zero-shot model. Second, we propose a relation prototype awareness module to make the zero-shot model aware of information contained in the prototypes. Last but not least, we develop an inheritance attention module to recalibrate the relation prototypes to enhance the inheritance process. The proposed framework is evaluated on two public cross-modality datasets including a cardiac dataset and an abdominal dataset. Extensive experiments show that the proposed framework significantly outperforms the state of the arts.
\end{abstract}

\begin{IEEEkeywords}
Zero-shot learning, Semantic segmentation, Multi-modality medical image.
\end{IEEEkeywords}

\section{Introduction}
\label{sec:introduction}

\begin{figure}[t]
	\centering
	\includegraphics[width=1.0\columnwidth]{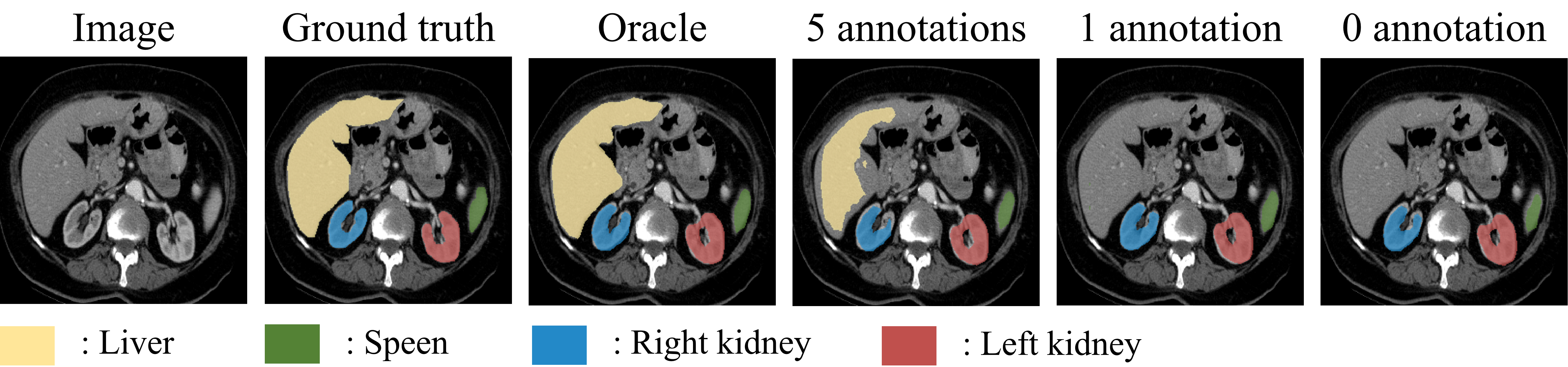}
	\caption{\small  Prediction results of liver by deep models with different quantities of annotations for training. Note that Oracle denotes the result of the model trained with full annotations (24 3D annotations in our case). }
	\label{fig:difficult_segmentation}
\end{figure}

\IEEEPARstart{O}{ver} the past decade, deep convolutional neural networks (DCNNs) have shown remarkable performance on various medical image processing tasks. A variety of frameworks based on DCNNs have been successfully applied to different medical image modalities and tasks~\cite{ronneberger2015u,wang2019patch,zhang2019toward,ouyang2020dual}.
Although great breakthroughs have been achieved for medical imaging, the current state-of-the-art (SOTA) performance is heavily data-driven by supervised learning.
To achieve reliable and robust performance, extensive manually labeled data are usually required, and when the annotated data become deficient or even absent, the model performance will drop dramatically as shown in Fig.~\ref{fig:difficult_segmentation}. Unfortunately, annotating large quantities of medical images is extremely laborious and expensive, especially for segmentation tasks whose goal is to classify each pixel in the image, as the delineation of organs and tissues in 3D medical images are challenging even for experienced physicians. Therefore, lack of well-annotated training data is a critical issue in medical image analysis.

To address the demanding need for labeled data, the key is to make full use of the available dataset. As a form of transfer learning, unsupervised domain adaptation (UDA)~\cite{long2015learning,tsai2018learning,chen2019synergistic} leverages the annotated source modality/domain data (\eg, magnetic resonance imaging (MRI)) to jointly train with the unlabeled target modality/domain data (\eg, computed tomography (CT)), to transfer the prior knowledge from the experienced source domain to the unfamiliar target domain. In this way, the shortage in labeled data is alleviated, since a well-trained model of the target domain can be replaced with a generalized model of the source domain. However, one limitation of the UDA methods is that, to transfer the prior knowledge, the source-domain data and labels must be available and utilized during the training process, which may be difficult to satisfy in many medical application scenarios due to data privacy concerns.

In contrast, zero-shot learning (ZSL)~\cite{lampert2009learning} has been actively investigated recently, which requires only the model trained in the source domain, therefore eliminating the requirement of labeled data from the source modality/domain. In ZSL, no labeled training images for a given set of test classes are required, yet we can still build an effective model by transferring knowledge from previously seen classes and certain auxiliary information. %Although it seems that ZSL and UDA are two separated approaches aiming at different tasks, these two methods share the same technical foundation--learning to project the source and target into a shared space.
{\color{blue}In most existing ZSL approaches for natural images~\cite{chen2018zero,huang2019generative,kato2019zero,min2020domain,wang2018zero,xian2019semantic,bucher2019zero}, the auxiliary information commonly refers to manual attributes~\cite{lampert2013attribute}, Word2Vec embedding~\cite{mikolov2013distributed} or WordNet lexical database~\cite{miller1995wordnet} which are utilized to form a vocabulary semantic space where the visual features of seen and unseen classes are projected and bridged. Different from the natural languages, however, the medical terminology is domain specific and professionally defined, causing a scarcity of work on ZSL in the medical field.}
Fortunately, multi-modality imaging, such as CT and MRI, is often used as a legitimate approach to acquire complementary information for better diagnosis. %Therefore, we set our target in the ZSL paradigm as to form the semantic space with annotated data from another medical imaging modality.
Therefore, we propose to leverage the information in data of an existing image modality with detailed annotations to generalize the visual semantics, and transfer this prior knowledge to the target task with a new image modality.

In this paper, we propose a novel annotation-efficient approach for medical image segmentation, where UDA is connected with ZSL in order to significantly reduce the workload of pixel-wise annotation for training DCNNs. Instead of using linguistic models to represent the semantic embedding space for bridging seen and unseen classes like most ZSL approaches do, we propose to utilize the prior knowledge about the segmentation targets, called relation prototypes, learned by a model from another image modality in a fully-supervised setting.
The proposed framework is based on the assumption that a prior segmentation model has been trained with a dataset of Modality A, where annotations of all classes are available. Then in the training stage with Modality B data, the well-trained prior model is utilized to provide the relation prototypes and guide the learning process of the zero-shot model with unannotated classes.

It is worth noting that our definition of ``unseen'' classes for cross-modal medical image segmentation is different from that in ZSL of natural images.
As multimodal medical images (\emph{e.g.}, CT and MRI) of the same body part usually capture consistent structures, ``unseen'' here actually means \textit{unannotated} classes in Modality B rather than strictly not seen before.
However, we still use this term for two reasons.
First, the unannotated classes are not perceived by the zero-shot model in Modality B and treated as background if not specially handled, which is the ``look but not see'' phenomenon.
Second, the concept of bridging the seen and unseen classes via semantic embedding is the same as the natural image ZSL in principle, despite the different sources of embedding (image and language).

{\color{blue}This notable difference makes existing ZSL methods developed for computer vision problems unable to be directly applied to the medical image field due to the large discrepancy of the problem definition between the two fields.
%However, it is not easy to directly follow or employ existing works in the medical field. The fundamental reason is that the scenario of the zero-shot learning in the computer vision for natural images could not apply in the medical field.
Specifically, ZSL with natural images chooses a set of objects as seen classes and excludes images of unseen classes from the training dataset.
However, in most medical images such as CT and MR, we cannot simply exclude certain structures as unseen classes, since normal human anatomical structures are unlikely to be absent in a healthy body.}
To this end, we formulate a new zero-shot semantic segmentation problem tailored for the medical scenario in this paper: a set of structures are annotated in one image modality and used to train a segmentation model, and the same structures are divided as seen (with annotations) and unseen (without annotation) classes in a new image modality.
The zero-shot segmentation algorithm is supposed to segment the unseen classes in the new image modality, given only: 1) the segmentation model pretrained in the first modality and 2) annotations of the seen classes in the new modality.
%Specifically, in natural images unseen classes can be excluded from the image content, i.e., remove the images that contain the unseen classes.
%However, we cannot simply exclude certain tissues in the medical image content, since human organs or anatomical structures are unlikely to be absent in a healthy body.
%To this end, we formulate a new zero-shot semantic segmentation problem tailored for the medical scenario in this paper.

\begin{figure}[t]
	\centering
    \vspace{1mm}
	\includegraphics[width=.95\columnwidth]{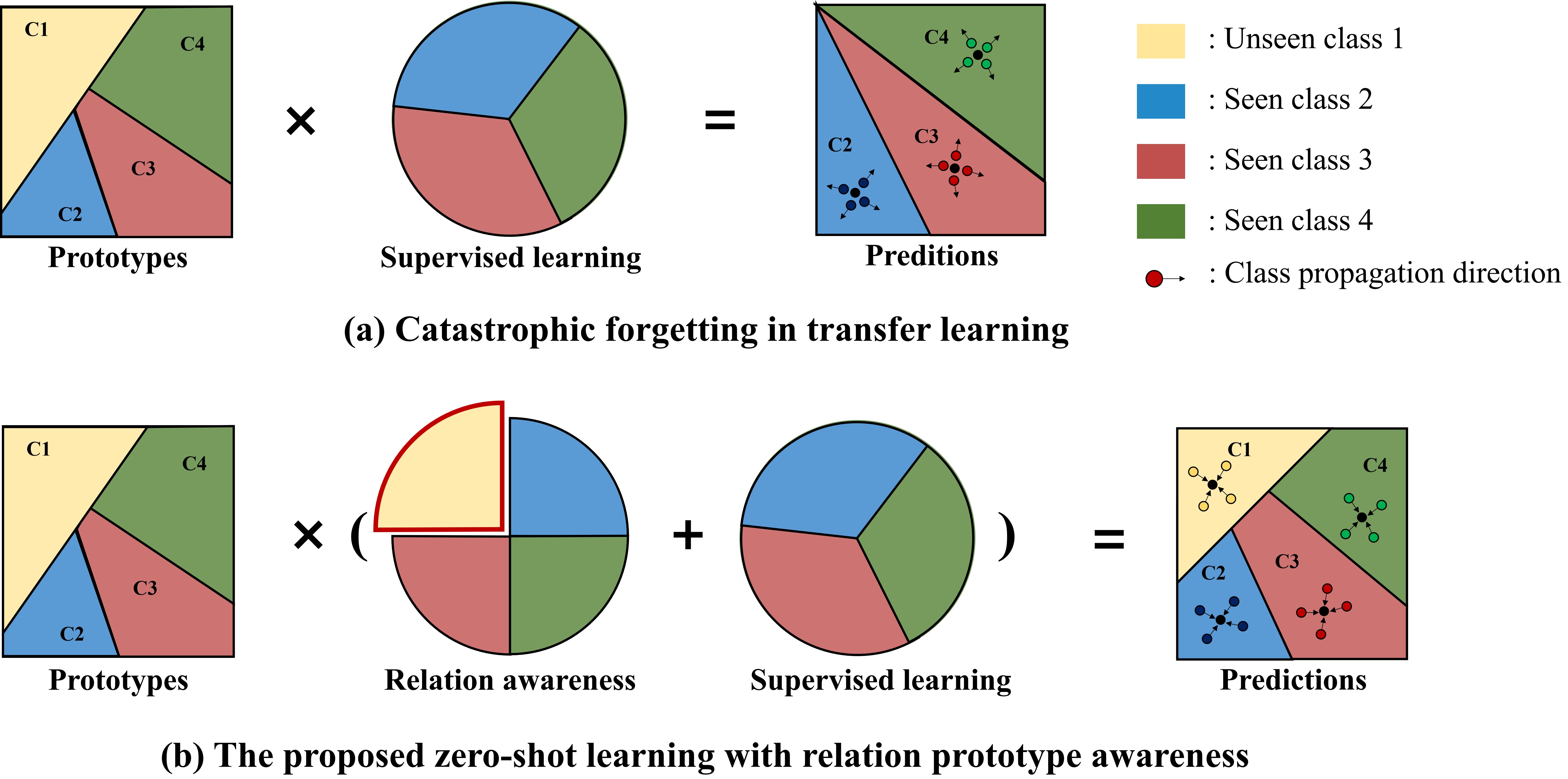}
	\caption{\color{blue}\small  An illustration of the catastrophic forgetting problem, and importance of the relation prototype awareness (RPA) for prevention. (a) Strong full supervision by seen classes will result in the memory loss on inherited features of unseen classes.
	(b)~The proposed RPA module maintains the function of the relation prototypes endowed with an anti-forgetting effect on unseen classes.}
	\label{fig:forget}
\end{figure}

{\color{blue}A unique catastrophic forgetting problem arises from the new problem setting of cross-modal zero-shot medical image segmentation.
As introduced above, the unseen classes are marked the same as the background in the annotations in Modality B.
As a result, the previously learned knowledge
%(\emph{e.g.}, the pre-trained weights of the zero-shot model)
by the prior model about the unseen classes will be forgotten under the strict supervision of seen classes as the training progresses (Fig.~\ref{fig:forget}(a)).}
%common problem in transfer learning
To address this problem, we propose to align the overall distributions of both the seen and unseen classes via adversarial training with a relation prototype awareness (RPA) module.
In this way,  the zero-shot model becomes aware of the features of the unseen classes and avoids overriding this critical part during the supervised training by the seen classes (Fig.~\ref{fig:forget}(b)).

In sum, UDA and ZSL are two useful approaches to the challenge of annotation absence, but their performances are conditioned on different prerequisites that cannot be easily fulfilled in the medical imaging field, \ie, UDA requires source-domain training data with full annotations and ZSL demands auxiliary information of rich domain description.
Our proposed method eliminates these obstacles by combining both approaches to move forwards true clinical utility. The core contributions of our framework are listed below:

\begin{itemize}
\item[1)] We propose a novel segmentation framework for medical images, which aims to improve the segmentation performance of unseen classes in a new modality by leveraging prior knowledge already learned in another modality.
    As far as the authors are aware of, this work is the first that exploits cross-modal image prior as the auxiliary information in ZSL, to make up the absence of linguistic models for medical terminology.
\item[2)] Compared to SOTA UDA approaches, our framework does not require any data but a well-trained model of the source modality, thus eliminating the problem of the source data privacy.
\item[3)] We propose a novel cross-modality adaptation (CMA) module to calibrate the common projection semantic space, enabling inheritance of the relation prototypes from the well-trained prior model to the zero-shot model.
\item[4)] To deal with the catastrophic forgetting problem in our context of cross-modal zero-shot segmentation---the zero-shot model forgets about unseen classes when trained with seen classes in a new modality, a relation prototype awareness (RPA) module is proposed to strengthen its memory of the unseen classes.
%learning process.
\item[5)] For better inheritance of the unseen classes, we design an inheritance attention (IA) module to recalibrate the features extracted by the zero-shot model.
\item[6)] Extensive experiments and analysis on two cross-modality datasets demonstrate the competency and efficacy of our framework.%is competent on the medical segmentation task.
\end{itemize}
{\color{blue}In conclusion, our work is a meaningful new application of the ZSL concept to an important new task. It features a novel problem setting for ZSL and several novel modules for an effective and integrative solution.}

\section{Related work}
\label{section:related work}
The main streams of ZSL methods can be roughly divided into the attribute-based~\cite{yu2010attribute,lampert2013attribute,parikh2011relative} and embedding-based methods~\cite{chen2018zero,huang2019generative,kato2019zero,min2020domain,wang2018zero,xian2019semantic,bucher2019zero,zhu2019semantic}. The attribute-based methods depend on feature engineering, where a semantic feature space is decomposed into a series of relative attribute spaces to generate a recognizable representation for unseen classes. Taking the task of abdominal multiple organ recognition as an example, if we set ${metabolic\; organ}$, ${regenerate\; organ}$ and ${coupled\; organ}$ to be the attributes, then the liver can be categorized by an attribute vector [1, 1, 0], whereas the kidney can be described by [1, 0, 1]. In contrast, the embedding-based methods learn the features of unseen classes from certain prior models' previously learned knowledge, where the relationship between the seen and unseen classes is already contained implicitly without the need for any manual engineering.

In this section, we first review previous standard approaches to ZSL. Then, a brief introduction of the UDA methods is given, followed by a discussion of the differences between UDA and ZSL.

\subsection{Attribute-based Zero-Shot Methods}
Attribute-based methods were commonly employed in earlier ZSL works.
Yu \etal\cite{yu2010attribute} proposed a generative attribute model to generate the object attributes and achieved satisfactory performance for ZSL.
Lampertet \etal\cite{lampert2013attribute} introduced multiple attribute classifiers for zero-shot classification.
The concept of relative attributes was proposed by Parikh \etal\cite{parikh2011relative}, which could associate the attributes with classes and indicate the relation levels, to further improve the zero-shot classification accuracy.
Although attribute-based methods substantively promoted the development of ZSL, constructing the unseen feature space with a set of discrete attribute spaces still heavily relies on manual design, lacking flexibility and scalability.
In contrast, embedding-based methods have gradually dominated in recent studies with the advancement of DCNNs.

\subsection{Embedding-based Zero-Shot Methods}
{\color{blue}The pioneering work by Blitzer \emph{et al.}~\cite{blitzer2009zero} established the theoretical grounding for embedding-based ZSL, with the general assumption that classes not/seldom seen could be learned via the correlation with a set of common seen classes shared between the domains/tasks.
Subsequently, this paradigm has been successfully applied to a wide variety of applications, such as pose estimation~\cite{kuznetsova2016exploiting}, video action ~\cite{fu2019embodied} and facial expression recognition \cite{wang2019comp}.}
%Prior knowledge about a set of seen and unseen classes (including their interrelations), called relation prototypes, can be learned from a well-trained prior model.
%Therefore, inferring unseen classes via inheriting the relation prototypes from the prior model is a feasible solution.
%%much easier than attribute representation.
As one of the most popular types of embedding for capturing such correlations, word embedding is a preferable and scalable option utilized in ZSL of computer vision, which was first proposed by Bengio \etal\cite{bengio2003neural} and extracted from a text corpus with a neural language model.

Successive advances of zero-shot research employed the word embedding method and achieved promising results~\cite{chen2018zero,huang2019generative,kato2019zero,min2020domain}. For instance, Wang \etal\cite{wang2018zero} improved the relation prototypes by combining knowledge graph and word embedding for better zero-shot recognition. Differently, Xian \etal\cite{xian2019semantic} proposed another method to achieve the inheritance of word embeddings by multiplying the framework's features with word embeddings directly. After that, Bucher \etal\cite{bucher2019zero} proposed a generative model for conditioning the inheritance of the word embeddings and visual features to boost the performance of zero-shot semantic segmentation.
Meanwhile,  Zhu \etal\cite{zhu2019semantic} proposed a semantic-guided multi-attention localization model for automatic joint learning of global and local features for ZSL.
However, all of these methods were based on word embeddings such as  visual descriptions,
%all these word embedding based methods are constructed upon a neural language prior model,
which are currently unavailable in the medical imaging field.
%Although Zhu\etal\cite{zhu2019semantic} proposed a viable solution to acquire relation prototypes by utilizing the global and local features extracted from the framework itself, these relation prototypes may not work well for medical data, since local features of medical images are not as discriminative as they are in the natural images.
To the best of our knowledge, this paper is the first to utilize image-based semantic embeddings for zero-shot segmentation of medical images.

In this work, we not only propose a new type of relation prototypes for zero-shot segmentation, but also propose a novel way to realize the inheritance of the relation prototypes, which is specially designed for the medical scenario.

\subsection{Unsupervised Domain Adaptation}
\label{section:related work_UDA}
As a subfield of transfer learning, UDA aims to transfer the well-tuned performance from one domain to another distinct domain. Recent UDA studies can be roughly divided into maximum discrepancy minimization~\cite{tzeng2014deep,long2015learning,sun2016deep,long2017deep}, adversarial-based distribution alignment~\cite{tzeng2017adversarial,kamnitsas2017unsupervised,ren2018adversarial,degel2018domain,bian2020uncertainty,chen2020generative} and image translation~\cite{zhu2017unpaired,zhang2018translating,zhang2018task,jiang2018tumor}. For the maximum discrepancy minimization methods, Tzeng \etal\cite{tzeng2014deep} proposed a maximum mean discrepancy (MMD) loss to minimize the domain shift between the source and target domains. Based on this work, Long \etal\cite{long2015learning,long2017deep} further revised the MMD loss with the proposed multiple kernel variant of maximum mean discrepancy (MK-MMD) loss and joint maximum mean discrepancy (JMMD) loss to achieve better UDA performance. With the development of the generative adversarial network (GAN), the adversarial-based distribution alignment and image translation methods have quickly dominated recent UDA research with superior performance.
For the adversarial-based distribution alignment methods, Tsai \etal~\cite{tsai2018learning} first proposed a domain adaptation module, where a discriminator was introduced to minimize the distance between distributions of the features from source and target domains, functioning as an MMD loss in effect. Luo \etal~\cite{luo2019taking} extended their work by introducing an auxiliary loss to align the local score map. Chang \etal~\cite{chang2019all} proposed a UDA solution combining disentangled representation learning and adversarial-based distribution alignment.
For the image translation methods, Chen \etal~\cite{chen2020unsupervised} merged the image translation using GAN and adversarial-based distribution alignment together, which worked effectively for the medical UDA problem. Li \etal~\cite{li2019bidirectional} brought forward the new concept of self-training based on image translation and verified the effectiveness of the proposed method.

However, a limitation of these UDA methods is that the source domain data are indispensable during the adaptation process, thus these methods are incapable to handle the medical application scenarios where the source training data are unavailable, \eg, due to privacy issue, when adapting to the target domain. Nevertheless, it is still valuable to compare with the UDA methods, since both the UDA and ZSL models would be trained without supervision in terms of the unannotated classes. Note that our approach requires annotating at least one class (\textit{i.e.}, organ or structure) of the target domain images as the seen class while treating other classes as unseen ones. If none of the annotation is available, the problem will degrade to UDA.

\section{Approach}
This section begins with the problem definition of the cross-modal annotation-efficient segmentation of medical images. Then, the details of proposed framework are elaborated. Last, an overall objective loss function is presented formally.

\subsection{Problem Definition}
\label{approach:problem_definition}
%This work aims to advance medical image zero-shot segmentation with the semantic embedding paradigm.

\begin{figure*}[t]
	\centering
	\includegraphics[width=0.95\textwidth]{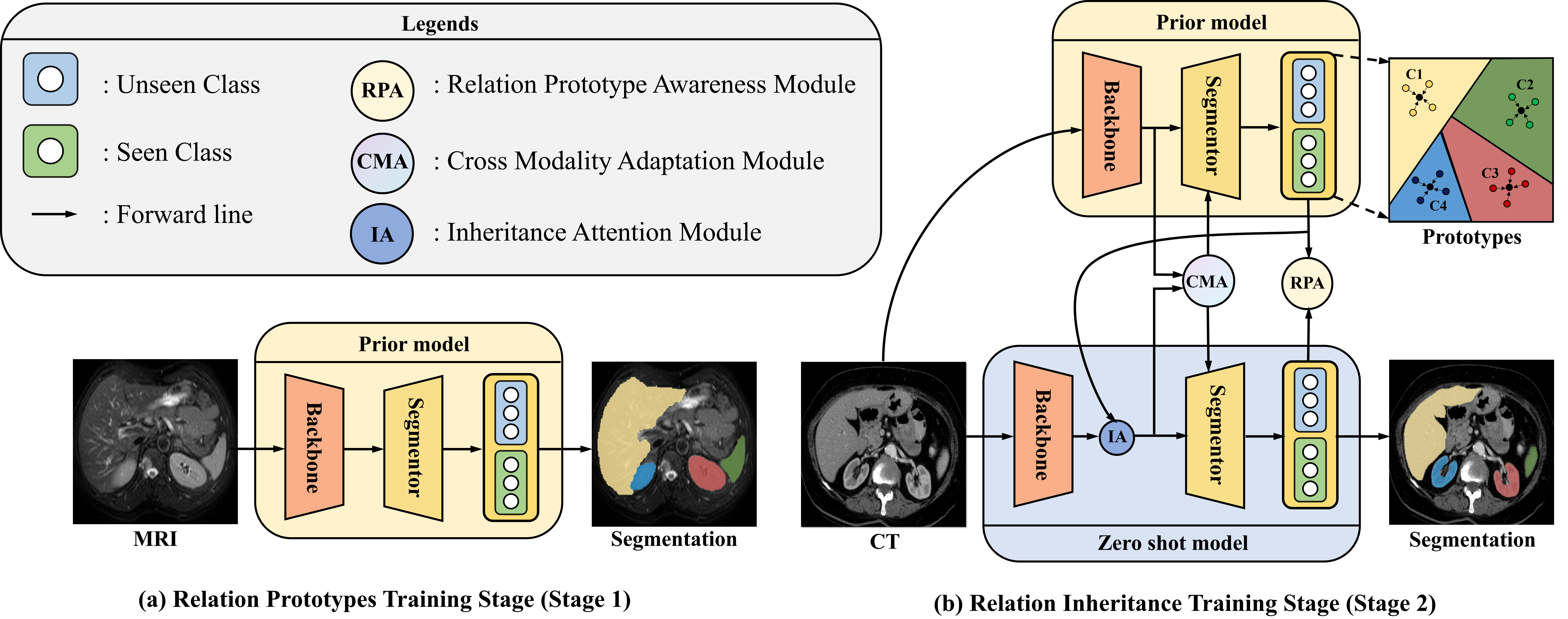}
	\caption{\small  Overview of the proposed zero-shot segmentation framework. Firstly, MRI images are used to train the prior model in Stage \textbf{1}. Then, in Stage \textbf{2}, the zero-shot model inherits the relation prototypes from the well-trained prior model via cross training with the cross-modality adaptation (CMA) module. The relation prototype awareness (RPA) module utilizes adversarial training and enables the zero-shot model to recognize unseen classes from the inherited relation prototypes. The inheritance attention (IA) module is proposed to improve the inheritance process.}
	\label{fig:overview}
\end{figure*}

Assume that a prior dataset with $C$ classes (including background) annotated is given for Modality A (\eg, MRI) with $N_p$ labeled training samples $\mathcal{D}_{p} = \{(x^{i}_{p}, y^{i}_{p})\}^{N_p}_{i=1}$, where $x^{i}_{p} \in \mathbb{R}^{W \times H \times 1}$ is the input image; $H$ and $W$ are the image height and width, respectively;
$y^{i}_{p} \in \{0,1\}^{W \times H \times C}$ is the corresponding segmentation mask, where $y_p^{i,w,h,c}\in\{0,1\}$ and is subject to $\sum_{c=1}^{C}{y_p^{i,w,h,c}}=1$, $\forall h \in \{1,...,H\}$, $\forall w \in \{1,...,W\}$.
Then, $\mathcal{D}_{p}$ is utilized to train a prior segmentation model, which can be any typical segmentation model such as DeepLabV3~\cite{chen2017rethinking}, in a fully supervised manner.
Once the prior training is done, the weights of the prior model are fixed and $\mathcal{D}_{p}$ are excluded from subsequent training procedures.
{\color{blue}The prior knowledge about each of the $C$ classes in $\mathcal{D}_p$ and their interrelations are learned and implicitly stored in the prior model, and are called the \textit{relation prototypes} in this work.}
Afterwards, we only need the well-trained prior model to extract relation prototypes for ZSL.

Suppose $\mathcal{D}_{s} = \{(x^{i}, y^{i}_{s})\}^{N_s}_{i=1}$ is the zero-shot segmentation training set, which contains $N_s$ samples related to the $C_{s}$ classes excluding background, where $x^{i} \in \mathbb{R}^{W \times H \times 1}$ is a training image of Modality B (\eg, CT) with the same structures as in $x_p$; $y^{i}_{s} \in \{0,1\}^{W \times H \times C_s}$ is the corresponding segmentation mask.
Note that the classes in $y_s$ are a strict subset of those in $y_p$, \emph{i.e.}, the \emph{seen} classes.
Let us denote segmentation masks of the \emph{unseen} classes (\emph{i.e.,} the classes in $y_p$ but not $y_s$) by $y_u\in\{0,1\}^{W\times H \times (C-C_s)}$.
%some of the classes in $y_p$ will be specified as \textit{unseen classes} $y_u$, which together with the background class will be assigned an invalid label. The {\color{red}classes} of the segmentation mask should satisfy: $y_s \cup y_u = y_p$ and $y_s \cap y_u = \varnothing$.
Our purpose is to effectively transfer the relation prototypes contained in the prior model to a zero-shot model, to help the latter produce accurate $y_u$ despite the absence of annotation for training.
We call this process the \textit{inheritance} of the relation prototypes.
For simplicity, we will drop the image index $i$ in the remainder of this paper if there is no ambiguity.

\subsection{Overview of the Proposed Framework}
In zero-shot segmentation methods for natural images, the relation prototypes are extracted with natural language models such as GloVe~\cite{pennington2014glove} and Word2Vec~\cite{mikolov2013distributed}.
Our method is fundamentally different in that, the relation prototypes are obtained from CNN models trained with images of a different modality.
The overall workflow of the proposed framework is shown in Fig.~\ref{fig:overview}, which includes the relation prototype training stage (Stage \textbf{1}) and inheritance training stage (Stage \textbf{2}).
In Stage~\textbf{1}, a prior model is trained with Modality A dataset $\mathcal{D}_{p}$ in a fully supervised manner. Once this training is completed, data from Modality A will not be used further. Correspondingly, the weights of the prior model are frozen.
Then, in Stage \textbf{2}, the zero-shot model is trained with the Modality B dataset $\mathcal{D}_{s}$ using the proposed relation prototype inheritance approach, comprising three modules:
(i) the cross-modality adaptation (CMA) module that makes use of the annotations of the seen classes to help the zero-shot model inherit the relation prototypes from the prior model, (ii) the relation prototype awareness (RPA) module that aligns the overall distributions of both the seen and unseen classes via adversarial training, and (iii) the inheritance attention (IA) module that recalibrates features of the zero-shot model based on the output of the prior model.
After Stage \textbf{2} training, the prototypes of unseen classes and their interrelations are transferred from the prior model to the zero-shot model and exploited by the zero-shot model's segmentor. %As a result, a comparable zero-shot segmentation is achieved.
More details about the training process are listed in Algorithm 1 and elaborated below.

\subsection{ Cross-Modality Adaptation}
\label{approach:Cross_accommodation_strategy}
%如前所述，模态A和模态B的数据集都是对应同一个结构不同模态的数据。我们假设这一种组织结构关系能够潜在被prototype所表达出来：也就是训练好的模态A的先验模型是有分割模态B中未见类别潜力的。因此，实现零样本分割的关键就是在stage 2中实现prototype的继承。
Given the problem setting in Section~\ref{approach:problem_definition}, both the Modality A and Modality B data capture the same anatomic structures, although in different modalities.
The topology of these structures is supposed to be represented in the relation prototypes implicitly, meaning that a well-trained prior model of Modality A has the potential to maintain the topology and segment unseen classes in Modality B when given annotations of the seen classes.
%the prior model trained with Modality A should be capable of recognizing identical structures and their relations in Modality B, including the unseen tissues, to some extent.}
Hence, the key to the proposed zero-shot semantic segmentation solution is the prototype inheritance in Stage \textbf{2}.
To achieve this goal, we propose a cross-modality adaptation (CMA) module.
Let $\mathcal{S}_{p}$ and $\mathcal{S}$ be segmentors of the prior model and the zero-shot model, respectively, where the weights of $\mathcal{S}_{p}$ are fixed during the entire Stage \textbf{2} training. As shown in Fig.~\ref{fig:overview}, we first feed the target image $x$ from Modality B to the feature extraction backbone networks of the two models to obtain the corresponding relation-prototype {\color{blue}feature map} $f_{p}$ from the prior model and the output feature map $f$ from the zero-shot model, respectively, {\color{blue}where $f_p, f\in\mathbb{R}^{W_f \times H_f \times C_f}$, $W_f$, $H_f$, and $C_f$ are the dimensions of the feature maps.}
Then we swap the obtained features and feed them to the segmentation networks $\mathcal{S}$ and $\mathcal{S}_{p}$ accordingly, to get the final outputs $m_{p\rightarrow s}=\mathcal{S}(f_{p})$ and $m_{s\rightarrow p}=\mathcal{S}_p(f)$. The difference between the outputs and ground truth can be formulated as the supervision loss $\mathcal{L}_{Cross}$:
%\begin{small}
%\begin{align*}
%    \label{eq:cross}
%    \mathcal{L}_{Cross}(x^i)=&-\frac{1}{K_s} \big(\sum_{k=1}^{K_s} y^{i}_{s}(k) %\log M^i_{p\rightarrow s}(k) \\
%    &+ \sum_{k=1}^{K_s} y_s^{i}(k) \log M^i_{s\rightarrow p}(k) \big),\tag{1}
%\end{align*}
%\end{small}
\begin{equation}
\begin{aligned}
    \label{eq:cross}
    \mathcal{L}_{Cross}(x)=&-\frac{1}{K_s} \Big[\sum_{k=1}^{K_s} y_{s}(k) \log m_{p\rightarrow s}(k) \\
    &+ \sum_{k=1}^{K_s} y_s(k) \log m_{s\rightarrow p}(k) \Big],
\end{aligned}
\end{equation}
%where $M_{p\rightarrow s}=\mathcal{S}(\mathcal{F}_{p})$ and $M_{s\rightarrow p}=\mathcal{S}_p(\mathcal{F})$ denote the output maps obtained by the segmentor with swapped features
where $k$ iterates over all locations and channels corresponding to the seen classes with $K_s = W\times H \times C_s$. Meanwhile, the zero-shot model should be trained in the fully supervised manner for the seen classes via $\mathcal{L}_{Seen}$:
\begin{equation}
\begin{aligned}
    \label{eq:seen}
    \mathcal{L}_{Seen}(x)= -\frac{1}{K_s}\sum_{k=1}^{K_s} y_s(k) \log m_s(k),
\end{aligned}
\end{equation}
where $m_s=\mathcal{S}(f)$.
%is the output map of the zero-shot model.

%Note that a special manual assignment is taken on $y_s$, as with mentioned in section \ref{approach:problem_definition},
\subsection{Awareness of Background}
Neither $\mathcal{L}_{Cross}$ nor $\mathcal{L}_{Seen}$ above takes into account the background class.
Several works~\cite{ravishankar2017joint,cermelli2020modeling} have shown the negative impact of neglecting the background on performance, as the background class may interfere the learning of target classes if not properly handled.
In our context, however, the zero-shot model has limited information about the background regions as it cannot distinguish the background from the unseen classes due to the absence of annotation.
%As reflected by the label annotation of the zero-shot training set, the unseen class is regarded the same as a member of background in $y_s$ rather than an independent class in $y_p$.
%For this reason, we allocate an invalid tag (\ie 255) to both background and unseen regions and ensure that they will not negatively affect on the backpropagation of $\mathcal{L}_{Cross}$ and $\mathcal{L}_{Seen}$.
%However, employing the aforementioned operation will inevitably neglect the special contribution of the background class and lead to bad performance. To address this problem,
Therefore, we utilize the pseudo label of the background class produced by the prior model to supervise the zero-shot model, so as to: (i) retrieve the missing information about the background regions, and (ii) reduce interference with the inheritance of the relation prototypes. Formally, the pseudo label supervised background awareness loss is defined as: %the self-training loss can be formulated as
\begin{equation}
\begin{aligned}
    \label{eq:bg}
    \mathcal{L}_{Bg}(x)&= -\frac{1}{K_{bg}}\sum_{k=1}^{K_{bg}} (\hat{y}_{bg}(k)- m_{s,bg}(k))^2, %\tag{3}
\end{aligned}
\end{equation}
where $\hat{y}_{bg} \in \{0,1\}^{W \times H \times 1}$ is the pseudo label of the background class generated by the prior model,
%is the background channel (\ie $c=0$) of the pseudo label $\hat{y} \in \{0,1\}^{W \times H \times C}$,
$m_{s,bg}$ is the background channel of $m_{s}$, and $k$ iterates over all locations of the background channel with $K_{bg}=W\times H$.

%After being processed by CMA, the relation prototype from the prior model will be completely inherited to the zero-shot model. In such a way, the model is endowed with the potential to identify unseen classes.

\subsection{Relation Prototype Awareness}
Although the relation prototypes of the seen classes can be effectively inherited from the prior model by the zero-shot model, the unseen classes are still unknown to the latter, since $\mathcal{S}$ is not supervised by any annotation of unseen classes.
As a result, the relation prototypes
%(\emph{i.e.}, the pretrained weights of the zero-shot model)
of the unseen classes will be forgotten under the strict supervision by seen classes (Eqs. (\ref{eq:cross}) and (\ref{eq:seen})), as illustrated in Fig.~\ref{fig:forget}(a).
%common problem during transfer learning: the relation prototype (\emph{e.g.}, the pre-trained weights of the zero-shot model) will be forgotten when directly using seen classes as the supervision for training.
The underlying reason is that the unseen classes have been marked the same as the background in the annotations, and their features are totally neglected.
As the training progresses, this issue will worsen and cause the catastrophic forgetting problem.
To address this problem, we propose a relation prototype awareness (RPA) module, which plays a crucial role in this work and enables the zero-shot model segmentor to recognize the relation prototypes of the unseen classes.
With the assistance of the proposed RPA module, the zero-shot model becomes aware of the features of the unseen class and avoids overriding this critical part during the supervised training by the seen classes, as illustrated in Fig.~\ref{fig:forget}(b).

%Given a fully-convolutional binary discriminator $\textbf{D}$, we then utilize it to classify output maps via the proposed loss, which can be formulated as:
% 正如我们提到的，导致灾难性遗忘现象的原因是由于可见类别强监督的错误引导。我们引入了对抗学习的模式来增强训练期间zero-shot 模型对prototype的记忆，缓解由于强监督损失造成的负面影响。具体来说，通过迭代训练的方式，即先冻结ZS模型训练D 4.a，然后冻结D，再利用对抗损失优化ZS模型 4.b， 来实现ZS模型对prototype的感知。

\begin{figure}[t]
	\centering
	\includegraphics[width=1.0\columnwidth]{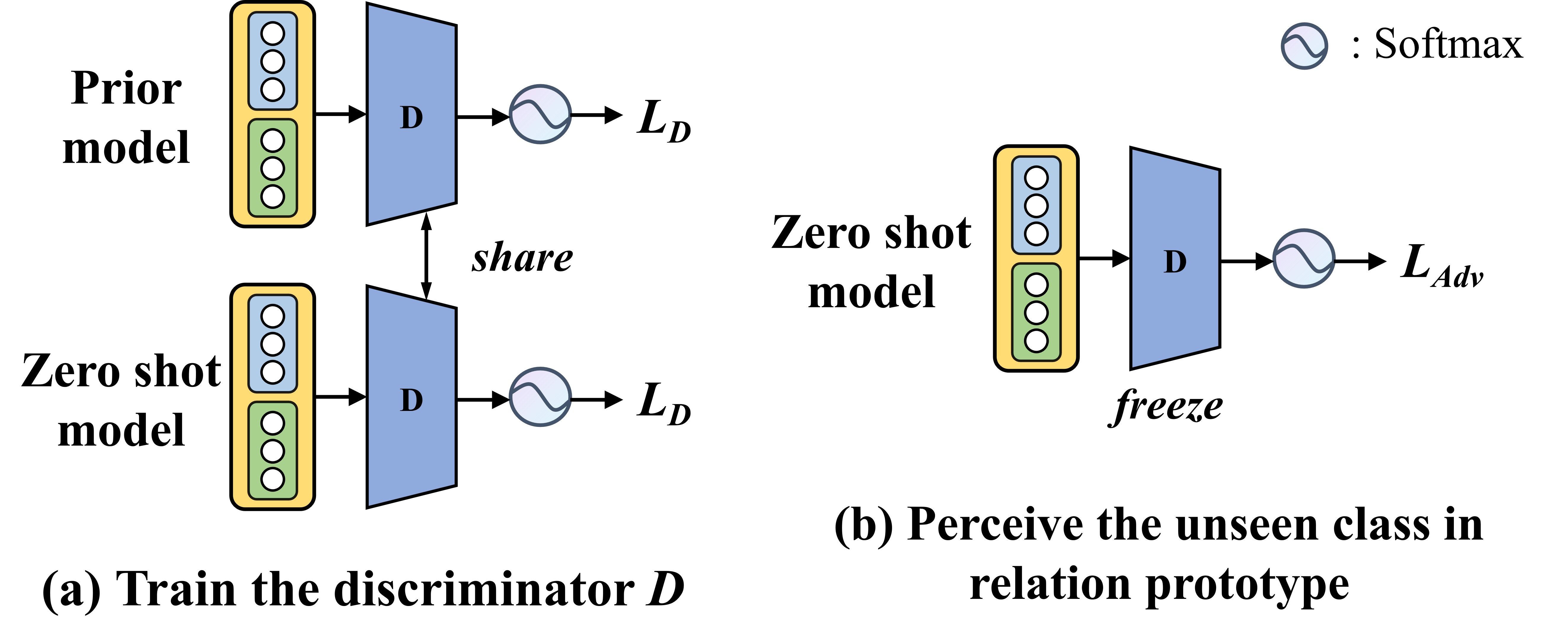}
	\caption{\small Structure of the relation prototype awareness module.}
	\label{fig:rpamodule}
\end{figure}

As shown in Fig.~\ref{fig:rpamodule}, the core component of RPA is a binary discriminator $\bm{D}$ implemented with PatchGAN~\cite{isola2017image} and composed of a series of convolutional and rectified linear unit (ReLU) layers.
Here, $m_p=\mathcal{S}_p(f_p)$ is considered as the positive sample (with the label 1), whereas $m_{s}$, $m_{s\rightarrow p}$, and $m_{p\rightarrow s}$ are considered as negative samples (with the label 0).
Then, the discrimination loss $\mathcal{L}_{D}$ is formulated as:
\begin{equation}\label{eq:L_D}
\begin{aligned}
    \mathcal{L}_{D}(x)&=-\lambda_{0}\log \bm{D}(m_{p})-\lambda_{1}\log [1-\bm{D}(m_{s\rightarrow p})]\\
                        &-\lambda_{2}\log [1-\bm{D}(m_{p\rightarrow s})]-\lambda_{3}\log [1-\bm{D}(m_{s})],
\end{aligned}
\end{equation}
where $\lambda_0, \lambda_1, \lambda_2, \lambda_3$ are hyper-parameters indicating the relative importance of the loss terms, and empirically set to 3, 1, 1, 1, respectively.
Since the catastrophic forgetting problem is due to the over-guidance of the seen classes' supervision, we introduce adversarial learning to consolidate the zero-shot model's memory of the unseen classes' prototypes and alleviate the negative effect caused by the strict supervision.
Specifically, we design an adversarial loss $\mathcal{L}_{Adv}$ to confuse $\bm{D}$ by pulling the distributions of $m_{s}$, $m_{s\rightarrow p}$, and $m_{p\rightarrow s}$ closer to that of $m_{p}$, which implicitly encourages the zero-shot model to produce output for the unseen classes that are indistinguishable from those by the prior model:
\begin{equation}
    \mathcal{L}_{Adv}(x)= -\log \bm{D}(m_{s\rightarrow p})-\log \bm{D}(m_{p\rightarrow s})-\log \bm{D}(m_{s}).
\end{equation}
In this way, the relation prototypes of the unseen classes can also be inherited from the prior model to the zero-shot model, overcoming the catastrophic forgetting problem.
It is worth noting that the goal of the RPA module is not to replicate $m_p$ precisely.
Instead, it helps the zero-shot model to learn about the unseen classes, via learning the distributions and interrelations of the seen and unseen classes that are captured by the prior model and implicitly embodied in $m_p$.
%by making the distributions of $m_s$, $m_{s\rightarrow p}$, and $m_{p\rightarrow s}$ indistinguishable from that of $m_p$

For implementation, the discriminator $\bm{D}$ and the zero-shot model are updated in an alternative manner. Specifically, we first freeze the zero-shot model to update $\bm{D}$ (Fig.~\ref{fig:rpamodule}(a); Step 11 in Algorithm 1). Then, $\bm{D}$ is frozen to update the zero-shot model (Fig.~\ref{fig:rpamodule}(b); Step 12 in Algorithm 1).% to achieve the awareness of the prototype

\begin{figure}[t]
	\centering
	\includegraphics[width=\columnwidth]{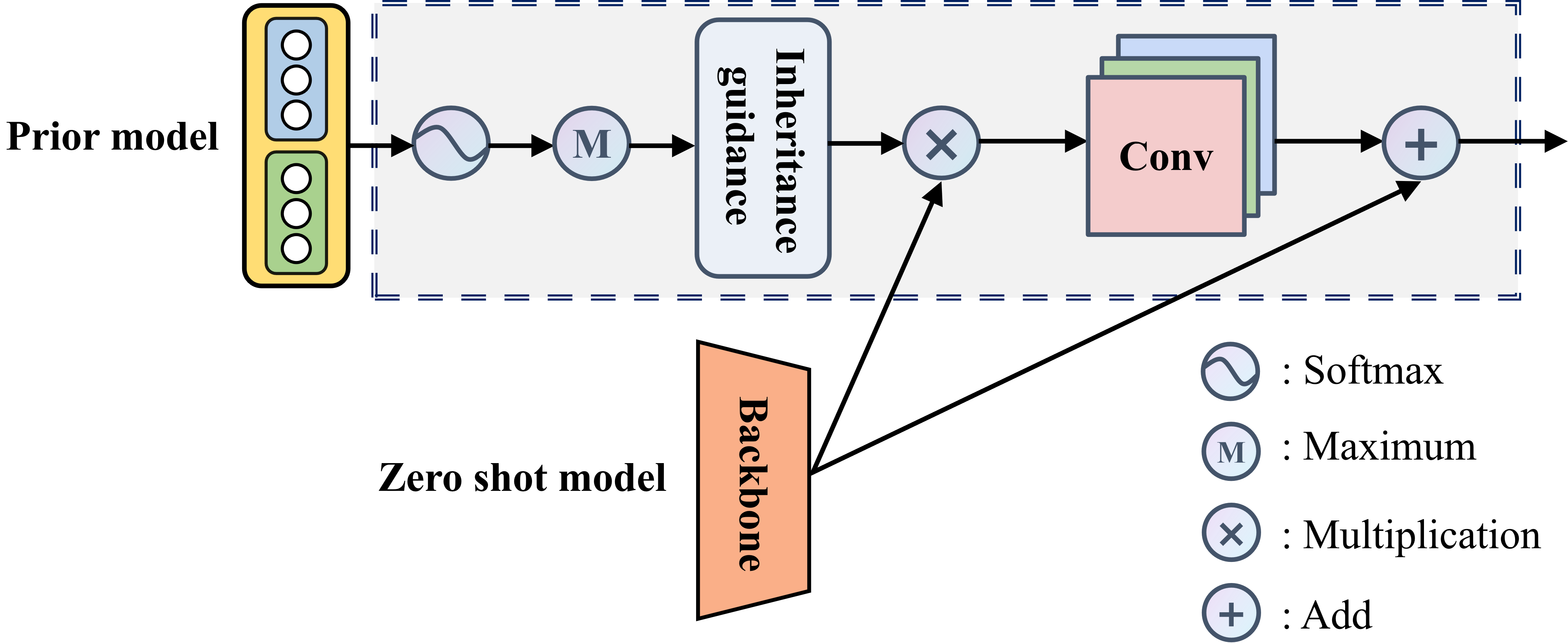}
	\caption{\small Structure of the inheritance attention module (enclosed in the dashed box).}
	\label{fig:iamodule}
\end{figure}

\subsection{Inheritance Attention}
Attention mechanism has proven effective in improving network performance by recalibrating features spatially and/or channel-wise~\cite{hu2018squeeze,fu2019dual,woo2018cbam}.
Being inspired, we form an inheritance attention (IA) module as shown in Fig.~\ref{fig:iamodule}.
Given that the relation prototypes are fully contained in the output $m_p$ of the prior model, the IA module generates an attentive guidance from $m_p$ for better inheritance.
%and utilize it across two models for better inheritance performance as shown in Fig.~\ref{fig:iamodule}. The proposed module is attached to the output of the prior model and merges features of the zero-shot model.
Specifically, we first feed $m_p$ into a softmax activation function $\sigma(\cdot)$ and calculate the maximum probability of all classes excluding the background to obtain an inheritance guidance map $g$:%since $M_{p}$ is the only clue that contains the relation prototype
\begin{equation}
\begin{aligned}
    \label{eq:guidance}
    g = \max\limits_{c}\{\sigma(m_{p})_c\},
\end{aligned}
\end{equation}
where $c$ indicates the channel number of a class other than background. %$c\neq0$ to exclude the background class
Then $g$ functions as an attention weight to enforce the zero-shot model to pay more attention to unseen classes:
\begin{equation}
\begin{aligned}
    \label{eq:attention}
    f_{s}=\bm{M}(g\otimes f)\oplus f,
\end{aligned}
\end{equation}
where $\bm{M}$ comprises several convolutional layers for feature fusion, $\otimes$ and $\oplus$ indicate the element-wise multiplication and summation, respectively, and $f_{s}$ is the zero-shot feature enhanced by the IA module.
Eventually, $f_s$ is used in place of $f$, where applicable.

\subsection{Overall Objective}
The overall objective function of the proposed framework is formulated as:
\begin{equation}
\begin{aligned}
    \label{eq:obj}
    \mathcal{L}_{Seg}(x)= \omega_{0}\mathcal{L}_{Cross}(x)+\omega_{1}\mathcal{L}_{Seen}(x&)\\+\omega_{2}\mathcal{L}_{Bg}(x)+\omega_{3}\mathcal{L}_{Adv}(x&),
\end{aligned}
\end{equation}
where $\omega_{0}$, $\omega_{1}$, $\omega_{2}$ and $\omega_{3}$ are the weights of the corresponding losses.
$\mathcal{L}_{Seg}$ and $\mathcal{L}_{D}$ (Eq. (\ref{eq:L_D})) are used to train the zero-shot model and discriminator $\bm{D}$, respectively, in an alternative manner.

\begin{algorithm}[t]
	\caption{Training procedure of the proposed framework.}
	\label{alg:Framwork}
	\small
	\begin{algorithmic}[1]
		\REQUIRE ~~\\
		Images $x_p$ and corresponding annotations $y_p$ from $\mathcal{D}_{p}$\\
		Images $x$ and annotations $y_{s}$ from $\mathcal{D}_{s}$\\
		\ENSURE ~~\\
		\textbf{Stage $1$:}\\
		\STATE Use $x_p$ and $y_p$ to train the prior model via the cross-entropy loss until convergence\\
		\textbf{Stage $2$:}\\
		\STATE Freeze the prior model weights
		\WHILE {$epoch \leq 250 $}
		\FOR {each sample $i$}
		\STATE Forward $x$ through the prior model, obtain $f_{p}$ and $m_{p}$
		\STATE Forward $x$ through backbone of the zero-shot model, obtain $f$
		\STATE Apply inheritance attention to $f$ based on $m_{p}$, obtain $f_{s}$
		\STATE Forward $f_s$ through the rest of the zero-shot model, obtain $m_{s}$
		\STATE Apply cross-modality adaptation to $f_{p}$ and $f_{s}$, obtain $m_{p\rightarrow s}$ and $m_{s\rightarrow p}$
        \STATE Compute $\mathcal{L}_{D}$ and $\mathcal{L}_{Seg}$ utilizing $m_{s}$, $m_{p}$, $m_{p\rightarrow s}$, and $m_{s\rightarrow p}$
		%\STATE Use {\color{red}$m_{s}$}, $m_{p}$, $m_{p\rightarrow s}$ and $m_{s\rightarrow p}$ to calculate $\mathcal{L}_{Cross}$, $\mathcal{L}_{Seen}$ and $\mathcal{L}_{Bg}$.
		%\STATE Unlock discriminator $\bm{D}$. Use {\color{red}$m_{s}$}, $m_{p}$, $m_{p\rightarrow s}$ and $m_{s\rightarrow p}$ to update $\bm{D}$ via $\mathcal{L}_{D}$.
        \STATE Update the discriminator $\bm{D}$ via $\nabla\mathcal{L}_{D}$, with the zero-shot model frozen
		%\STATE Freeze $\bm{D}$, and utilize {\color{red}$m_{s}$}, $m_{p}$, $m_{p\rightarrow s}$ and $m_{s\rightarrow p}$ to calculate $\mathcal{L}_{Adv}$.
		%\STATE Calculate $\mathcal{L}_{Seg}$, and back-propagate to update the zero-shot model.
        \STATE Update the zero-shot model via $\nabla\mathcal{L}_{Seg}$, with the discriminator $\bm{D}$ frozen
				
		\ENDFOR

		\ENDWHILE
	\end{algorithmic}
\end{algorithm}

% \begin{figure*}[bt]
% 	\centering
% 	\includegraphics[width=0.7\textwidth]{Fig/structure0120.jpg}
% 	\caption{\small  Visual examples of the ablation study results on the abdominal dataset. The color block on the bottom-left corner of the image means the corresponding structure is specified as the unseen class. As can be seen, the proposed modules can effectively improve the segmentation of unseen classes.
% 	}
% 	\label{fig:ablation_abdominal}
% \end{figure*}

\section{Experiments}
\subsection{Datasets}
Two cross-modality datasets are used in this study for evaluation, including an abdominal dataset and a cardiac dataset.
For the abdominal dataset, we obtain 20 MRI volumes from the CHAOS Challenge \cite{kavur2020chaos} and 30 CT volumes from \cite{landman2015multi}, respectively.
Manual delineations involve multiple organs, including liver, right kidney (R-Kid), left kidney (L-Kid), and spleen.
The cardiac dataset is collected from the \textit{MMWHS} challenge~\cite{zhuang2016multi}, including 20 MRI volumes %(with the valid slices of 47 to 127)
and 20 CT volumes. %(ranging from 142 to 251 valid slices)
Four anatomical structures are annotated for both modalities: the ascending aorta (AA), the left atrium blood cavity (LAB), the left ventricle blood cavity (LVB), and the myocardium of the left ventricle (MYO).
Note that there is no cross-body-part experiment (\textit{i.e.}, the two datasets are used separately), as there usually exists no structural correspondence in such setting.

%To simplify our experiments,
Following the convention in the UDA literature~\cite{chen2019synergistic,bian2020uncertainty}, the MRI modality in both datasets is used as the labeled prior set $\mathcal{D}_{p}$ to train the prior model.
%only but not the zero-shot model.
The CT modality is used as the zero-shot dataset $\mathcal{D}_{s}$ to train the zero-shot model.
For the prior model, all scans in $\mathcal{D}_{p}$ are used for training.
As to the zero-shot model, scans in $\mathcal{D}_{s}$ are randomly divided into 80\% for training and 20\% for testing, respectively.
The selection of the unseen classes and the influence will be discussed later. %It is worth noting that the unseen classes will be provided in the annotation of the testing scan.

Since the two modalities in each of the two datasets are collected from different clinical centers, their views are not perfectly matched.
To eliminate potential influence, we follow Chen \etal \cite{chen2020unsupervised} to roughly align the two modalities by manually cropping them to the same view and removing non-informative slices.
Specifically, for the abdominal dataset we discard the axial slices that do not contain any of the four target organs and crop out the non-body region, whereas for the cardiac dataset, we cut out a 3D volume that is 1.25 times the bounding box of the heart.
%The numbers of slices before and after this step are tabulated in Table~\ref{table:dataset}.
More details are shown in Table~\ref{table:dataset}.
After that, we resize the image slices to 256$\times$256 pixels.
Standard normalization has been performed on both datasets by subtracting the mean value and then divided by the standard deviation. Data augmentation procedures such as random flip, rotation, and scaling are adopted to reduce overfitting.

\subsection{Experimental Configuration}
The experimental configuration is designed to evaluate the effectiveness of the proposed zero-shot segmentation framework. Most recent ZSL studies mainly focus on improving the performance on unseen classes but neglect to evaluate the performance of seen classes. On the contrary, our experiments involve a comprehensive evaluation covering both seen and unseen classes. All experiments are performed on both of the abdominal and cardiac datasets. The detailed settings are:
\begin{itemize}	
\item[a)] To evaluate the effectiveness of the proposed modules, including RPA, CMA, and IA, module ablations are conducted treating each cardiac structure/abdominal organ as the unseen class.
\item[b)] To evaluate the robustness against the number of unseen classes, ablation studies on different combinations of unseen classes on both datasets are conducted.
\item[c)] To evaluate the final segmentation performance, comparisons against several SOTA UDA and ZSL methods~\cite{tsai2018learning,luo2019taking,li2019bidirectional,chang2019all,chen2020unsupervised,bian2020uncertainty,bucher2019zero} are made. For a fair comparison,
%{\color{red} with the UDA methods where all classes will be regarded as unseen {\color{red}classes}}
the experiment is conducted repeatedly with our proposed method by taking each class as the unseen class, and the results for each unseen class are reported.
\end{itemize}

\renewcommand{\arraystretch}{1}
\begin{table}[t]
	\centering
	\Large
	\caption{Statistics of the cross-modality abdominal multi-organ~\cite{kavur2020chaos} and cardiac~\cite{zhuang2016multi} datasets.}
	\label{table:dataset}
	\scalebox{0.5}{
	\begin{tabular}{p{2.2cm}<{\centering}|p{1.6cm}<{\centering}|p{1.6cm}<{\centering}|p{1.1cm}<{\centering}|p{1.9cm}<{\centering}|p{1.9cm}<{\centering}|p{1.9cm}<{\centering}|p{1.9cm}<{\centering}}
		\toprule[1pt]
		\toprule[1pt]
		\multirow{2}{*}{Dataset}&\multicolumn{1}{c|}{\multirow{2}{*}{Modality}}&\multicolumn{1}{c|}{\multirow{2}{*}{Patients}}&\multicolumn{1}{c|}{\multirow{2}{*}{Role}}&\multicolumn{2}{c|}{Image size}&\multicolumn{2}{c}{Slice number}\\
		\cline{5-8}
		\multicolumn{1}{c|}{} & \multicolumn{1}{c|}{} & \multicolumn{1}{c|}{} &\multicolumn{1}{c|}{} & \multirow{1}{*}[0ex]{Original} & \multirow{1}{*}[0ex]{Processed} & \multirow{1}{*}[0ex]{Original} &\multirow{1}{*}[0ex]{Processed} \\
		
		%\cline{9-10} & Original&Processd & \multicolumn{2}{c|}{}\\

		\toprule[1pt]
		\bottomrule[1pt]
		\multirow{2}{*}{Abdominal}& \multirow{1}{*}[0ex]{MRI} &20  &$\mathcal{D}_{p}$&256 &256 &26$\sim$39 &20$\sim$30 \\
		&CT  &30 &$\mathcal{D}_{s}$ &512 & 256   &58$\sim$198 & 35$\sim$117 \\
		\hline
		\multirow{2}{*}{Cardiac}&MRI & 20 &$\mathcal{D}_{p}$ &256$\sim$512 & 256  &112$\sim$200 &47$\sim$127\\
		&CT & 20 &$\mathcal{D}_{s}$ & 512 &  256  & 177$\sim$363 &142$\sim$251\\
        \bottomrule[1pt]
		\bottomrule[1pt]
	\end{tabular}}
\end{table}

In addition, a \textit{lower-bound} model and an \textit{Oracle} model, which are frequently used in most UDA studies as benchmarks representing the lower and upper bounds of the performance, are included. Suppose we have a fully supervised ``mirror'' dataset of $\mathcal{D}_s$, denoted as $\mathcal{D}_m$, in which all classes are annotated for training. Then, the lower-bound model is trained with $\mathcal{D}_{p}$ and tested on $\mathcal{D}_{m}$, while Oracle is trained with $\mathcal{D}_{m}$ and tested on $\mathcal{D}_{m}$.

\begin{table*}[ht]
    \caption{Ablation studies on the proposed modules. Shaded blocks indicate the unseen class in the current experiment. Dice score is used as the evaluation metric and reported in the percentage (\%).}
    \label{table:module_ablations}
	\centering
	\renewcommand{\arraystretch}{0.9}
	\scalebox{0.46}{
	\LARGE
	\begin{tabular}{C{7.0cm}|C{2.0cm}|C{1.5cm}|C{1.5cm}|C{1.5cm}|C{1.5cm}|C{1.5cm}|C{1.5cm}|C{1.5cm}|C{1.5cm}|C{1.5cm}|C{1.5cm}|C{1.5cm}|C{1.5cm}|C{1.5cm}|C{1.5cm}}
    	\toprule[1pt]
		\toprule[1pt]
        \multirow{3}{*}{\bf{Unseen class}}&\multicolumn{5}{C{10.0cm}|}{\multirow{1}{*}{\shortstack{\bf{Ablation}}}}
 		& \multicolumn{5}{C{9.0cm}|}{\bf{Abdominal}}& \multicolumn{5}{C{8.5cm}}{\bf{Cardiac}}\\
 		\cline{2-16}
 		&\multicolumn{1}{C{2.0cm}|}{\multirow{2}{*}{\shortstack{\bf{Setting}}}}
 		&\multicolumn{1}{C{1.5cm}|}{\multirow{2}{*}{\shortstack{\bf{$\mathcal{L}_{Bg}$}}}}
 		&\multicolumn{1}{C{1.5cm}|}{\multirow{2}{*}{\shortstack{\bf{RPA}}}}
 		&\multicolumn{1}{C{1.5cm}|}{\multirow{2}{*}{\shortstack{\bf{CMA}}}}
 	    &\multicolumn{1}{C{1.5cm}|}{\multirow{2}{*}{\shortstack{\bf{IA}}}}
 		&\multicolumn{1}{C{1.5cm}|}{\multirow{2}{*}{\shortstack{\bf{Liver}}}}
    	&\multicolumn{1}{C{1.5cm}|}{\multirow{2}{*}{\shortstack{\bf{R-Kid}}}}
 		&\multicolumn{1}{C{1.5cm}|}{\multirow{2}{*}{\shortstack{\bf{L-Kid}}}}
 		&\multicolumn{1}{C{1.5cm}|}{\multirow{2}{*}{\shortstack{\bf{Speen}}}}
 		&\multicolumn{1}{C{1.5cm}|}{\multirow{2}{*}{\shortstack{\bf{Mean}}}}
 		&\multicolumn{1}{C{1.5cm}|}{\multirow{2}{*}{\shortstack{\bf{AA}}}}
    	&\multicolumn{1}{C{1.5cm}|}{\multirow{2}{*}{\shortstack{\bf{LAB}}}}
 		&\multicolumn{1}{C{1.5cm}|}{\multirow{2}{*}{\shortstack{\bf{LVB}}}}
 		&\multicolumn{1}{C{1.5cm}|}{\multirow{2}{*}{\shortstack{\bf{MYO}}}}
 		&\multicolumn{1}{C{1.5cm}}{\multirow{2}{*}{\shortstack{\bf{Mean}}}}\\
 		\multirow{7}{*}{\shortstack{\\\\\\\\\\\\\bf{Liver/AA}}}
 	    &\quad&\quad&\quad&\quad&\quad&\quad&\quad&\quad&\quad&\quad&\quad&\quad&\quad&\quad\\
 	    \toprule[1pt]
 	    \bottomrule[1pt]
 	    &(a)\quad&\multicolumn{4}{c|}{Baseline}&38.06\cellcolor[rgb]{.7,.9,.9}&15.19\quad&20.25\quad&21.22\quad&23.68\quad&2.03\cellcolor[rgb]{.7,.9,.9}&60.19\quad&72.37\quad&28.99\quad&40.89\quad\\
 	    \cline{3-6}
 	    &(b)\quad&\quad&\quad&\quad&\quad&1.18\cellcolor[rgb]{.7,.9,.9}&6.21\quad&11.11\quad&7.69\quad&6.58\quad&00.48\cellcolor[rgb]{.7,.9,.9}&20.79\quad&72.86\quad&12.46\quad&26.65\quad\\ % baseline
 	    &(c)\quad&\checkmark&\quad&\quad&\quad&00.00\cellcolor[rgb]{.7,.9,.9}&64.98\quad&70.79\quad&78.84\quad&53.65\quad&00.00\cellcolor[rgb]{.7,.9,.9}&84.13\quad&88.65\quad&62.03\quad&58.70\quad\\ % Lbg
 	     %&(d)\quad&\checkmark&\checkmark&\quad&\quad&77.38\cellcolor[rgb]{.7,.9,.9}&82.92\quad&80.77\quad&82.76\quad&80.96\quad&63.31\cellcolor[rgb]{.7,.9,.9}&87.68\quad&83.28\quad&74.87\quad&72.82\quad\\   %Lbg+RPA
 	     %&(a)\quad&\checkmark&\quad&\checkmark&\quad&00.00\cellcolor[rgb]{.7,.9,.9}&77.09\quad&76.82\quad&76.07\quad&57.50\quad&00.00\cellcolor[rgb]{.7,.9,.9}&87.93\quad&88.65\quad&77.16\quad&63.43\quad\\ %Lbg+CMA
 	     %&(a)\quad&\checkmark&\quad&\quad&\checkmark&00.00\cellcolor[rgb]{.7,.9,.9}&71.59\quad&77.21\quad&77.87\quad&56.68\quad&00.00\cellcolor[rgb]{.7,.9,.9}&90.10\quad&89.05\quad&70.64\quad&62.45\quad\\ %Lbg+IA
 	    &(d)\quad&\checkmark&\checkmark&\checkmark&\quad&85.55\cellcolor[rgb]{.7,.9,.9}&83.39\quad&82.73\quad&86.20\quad&84.47\quad&73.09\cellcolor[rgb]{.7,.9,.9}&91.34\quad&86.76\quad&78.67\quad&82.47\quad\\ %Lbg+RPA+CMA
   	    &(e)\quad&\checkmark&\checkmark&\quad&\checkmark\quad&81.26\cellcolor[rgb]{.7,.9,.9}&80.41\quad&81.44\quad&86.72\quad&82.46\quad&75.49\cellcolor[rgb]{.7,.9,.9}&92.20\quad&87.34\quad&75.88\quad&82.73\quad\\ %Lbg+RPA+IA
   	    &(f)\quad&\checkmark&\quad&\checkmark&\checkmark&00.00\cellcolor[rgb]{.7,.9,.9}&80.83\quad&83.16\quad&77.57\quad&60.39\quad&00.00\cellcolor[rgb]{.7,.9,.9}&89.49\quad&89.75\quad&79.51\quad&64.69\quad\\ % Lbg+CMA+IA
 	    &(g)\quad&\checkmark&\checkmark&\checkmark&\checkmark&90.61\cellcolor[rgb]{.7,.9,.9}&84.15\quad&86.08\quad&88.95\quad&87.45\quad&84.66\cellcolor[rgb]{.7,.9,.9}&91.63\quad&89.20\quad&80.92\quad&86.60\quad\\ % all
 	    \cline{1-16}
 	    \multirow{7}{*}{\shortstack{\\\\\\\\\\\bf{R-Kid/LAB}}}
 	    &(a)\quad&\multicolumn{4}{c|}{Baseline}&52.60\quad&00.10\cellcolor[rgb]{.7,.9,.9}&24.56\quad&28.09\quad&26.34\quad&43.79\quad&1.67\cellcolor[rgb]{.7,.9,.9}&57.93\quad&29.84\quad&33.31\quad\\
 	    \cline{3-6}
        &(b)\quad&\quad&\quad&\quad&\quad&32.70&00.00\cellcolor[rgb]{.7,.9,.9}\quad&11.67\quad&11.68\quad&14.01\quad&16.64&00.00\cellcolor[rgb]{.7,.9,.9}\quad&74.88\quad&12.22\quad&25.94\quad\\
 	    &(c)\quad&\checkmark&\quad&\quad&\quad&85.68&00.00\cellcolor[rgb]{.7,.9,.9}\quad&74.38\quad&81.79\quad&60.46\quad&85.59&00.00\cellcolor[rgb]{.7,.9,.9}\quad&84.64\quad&66.25\quad&59.12\quad\\
 	    %&(a)\quad&\checkmark&\checkmark&\quad&\quad&89.47\quad&62.34\cellcolor[rgb]{.7,.9,.9}&80.28\quad&80.75\quad&78.21\quad&88.85\quad&78.41\cellcolor[rgb]{.7,.9,.9}&83.51\quad&78.91\quad&82.67\quad\\
 	    %&(a)\quad&\checkmark&\quad&\checkmark&\quad&90.91&00.00\cellcolor[rgb]{.7,.9,.9}\quad&81.11\quad&79.78\quad&62.95\quad&83.45&00.00\cellcolor[rgb]{.7,.9,.9}\quad&88.97\quad&74.72\quad&61.78\quad\\
 	    %&(a)\quad&\checkmark&\quad&\quad&\checkmark&89.56&00.00\cellcolor[rgb]{.7,.9,.9}\quad&77.15\quad&77.62\quad&61.08\quad&81.03&00.00\cellcolor[rgb]{.7,.9,.9}\quad&88.76\quad&74.87\quad&61.16\quad\\
 	    &(d)\quad&\checkmark&\checkmark&\checkmark&\quad&91.65\quad&76.15\cellcolor[rgb]{.7,.9,.9}&81.95\quad&85.10\quad&83.71\quad&89.64\quad&81.42\cellcolor[rgb]{.7,.9,.9}&86.86\quad&79.51\quad&84.36\quad\\
 	    &(e)\quad&\checkmark&\checkmark&\quad&\checkmark&91.44&75.91\cellcolor[rgb]{.7,.9,.9}\quad&81.39\quad&83.47\quad&83.05\quad&93.04&82.78\cellcolor[rgb]{.7,.9,.9}\quad&86.96\quad&77.45\quad&85.06\quad\\
 	    &(f)\quad&\checkmark&\quad&\checkmark&\checkmark&88.71&00.00\cellcolor[rgb]{.7,.9,.9}\quad&79.44\quad&77.31\quad&61.36\quad&87.67&00.00\cellcolor[rgb]{.7,.9,.9}\quad&88.30\quad&79.85\quad&63.95\quad\\
 	     &(g)\quad&\checkmark&\checkmark&\checkmark&\checkmark&92.44\quad&82.09\cellcolor[rgb]{.7,.9,.9}&83.45\quad&87.53\quad&86.38\quad&93.06\quad&87.55\cellcolor[rgb]{.7,.9,.9}&88.64\quad&79.34\quad&87.15\quad\\
 	    \cline{1-16}
 	    \multirow{7}{*}{\shortstack{\\\\\\\\\\\bf{L-Kid/LVB}}}
 	    &(a)\quad&\multicolumn{4}{c|}{Baseline}&56.02\quad&31.13\quad&00.00\cellcolor[rgb]{.7,.9,.9}&19.70\quad&26.71\quad&49.46\quad&62.13\quad&00.00\cellcolor[rgb]{.7,.9,.9}&31.06\quad&35.66\quad\\
 	    \cline{3-6}
 	    &(b)\quad&\quad&\quad&\quad&\quad&36.12\quad&12.70\quad&00.00\cellcolor[rgb]{.7,.9,.9}&8.92\quad&14.43\quad&17.67\quad&52.54\quad&00.00\cellcolor[rgb]{.7,.9,.9}&14.28\quad&21.12\quad\\
 	    &(c)\quad&\checkmark&\quad&\quad&\quad&85.46\quad&71.85\quad&00.00\cellcolor[rgb]{.7,.9,.9}&77.32\quad&58.66\quad&78.72\quad&90.54\quad&00.00\cellcolor[rgb]{.7,.9,.9}&65.95\quad&58.80\quad\\
 	    %&(d)\quad&\checkmark&\checkmark&\quad&\quad&87.23\quad&79.64\quad&63.24\cellcolor[rgb]{.7,.9,.9}&80.27\quad&77.60\quad&86.28\quad&89.76\quad&75.99\cellcolor[rgb]{.7,.9,.9}&69.57\quad&80.40\quad\\
 	    %&(a)\quad&\checkmark&\quad&\checkmark&\quad&87.94\quad&80.22\quad&00.00\cellcolor[rgb]{.7,.9,.9}&81.91\quad&62.52\quad&84.97\quad&92.68\quad&00.00\cellcolor[rgb]{.7,.9,.9}&78.53\quad&64.04\quad\\
 	    %&(a)\quad&\checkmark&\quad&\quad&\checkmark&89.85\quad&78.66\quad&00.00\cellcolor[rgb]{.7,.9,.9}&77.36\quad&61.47\quad&81.70\quad&89.92\quad&00.00\cellcolor[rgb]{.7,.9,.9}&71.51\quad&60.78\quad\\
 	    &(d)\quad&\checkmark&\checkmark&\checkmark&\quad&91.60\quad&81.67\quad&76.78\cellcolor[rgb]{.7,.9,.9}&84.43\quad&83.62\quad&89.87\quad&89.05\quad&82.56\cellcolor[rgb]{.7,.9,.9}&76.10\quad&84.40\quad\\
 	    &(e)\quad&\checkmark&\checkmark&\quad&\checkmark&92.88\quad&82.88\quad&72.24\cellcolor[rgb]{.7,.9,.9}&87.09\quad&83.77\quad&86.89\quad&87.69\quad&82.99\cellcolor[rgb]{.7,.9,.9}&73.60\quad&82.81\quad\\
 	    &(f)\quad&\checkmark&\quad&\checkmark&\checkmark&90.67\quad&83.61\quad&00.00\cellcolor[rgb]{.7,.9,.9}&82.18\quad&64.12\quad&88.27\quad&93.55\quad&00.00\cellcolor[rgb]{.7,.9,.9}&78.89\quad&65.18\quad\\
 	    &(g)\quad&\checkmark&\checkmark&\checkmark&\checkmark&93.50\quad&84.52\quad&82.33\cellcolor[rgb]{.7,.9,.9}&89.02\quad&87.34\quad&92.74\quad&92.66\quad&86.92\cellcolor[rgb]{.7,.9,.9}&79.52\quad&87.96\quad\\
 	    \cline{1-16}
 	    \multirow{7}{*}{\shortstack{\\\\\\\\\\\bf{Spleen/MYO}}}
 	     &(a)\quad&\multicolumn{4}{c|}{Baseline}&53.68\quad&30.23\quad&18.04\quad&00.36\cellcolor[rgb]{.7,.9,.9}&25.58\quad&46.15\quad&44.98\quad&46.13\quad&00.08\cellcolor[rgb]{.7,.9,.9}&34.33\quad\\
 	    \cline{3-6}
 	    &(b)\quad&\quad&\quad&\quad&\quad&34.00\quad&15.34\quad&6.83\quad&00.00\cellcolor[rgb]{.7,.9,.9}&14.04\quad&16.39\quad&34.05\quad&15.51\quad&00.00\cellcolor[rgb]{.7,.9,.9}&16.49\quad\\
 	    &(c)\quad&\checkmark&\quad&\quad&\quad&86.61\quad&72.23\quad&73.60\quad&00.00\cellcolor[rgb]{.7,.9,.9}&58.11\quad&87.69\quad&85.69\quad&77.31\quad&00.00\cellcolor[rgb]{.7,.9,.9}&62.67\quad\\
 	    %&(d)\quad&\checkmark&\checkmark&\quad&\quad&90.57\quad&80.03\quad&79.50\quad&67.39\cellcolor[rgb]{.7,.9,.9}&79.37\quad&85.77\quad&84.20\quad&83.27\quad&53.88\cellcolor[rgb]{.7,.9,.9}&76.78\quad\\
 	    %&\checkmark&\quad&\checkmark&\quad&83.88\quad&74.57\quad&70.96\quad&00.00\cellcolor[rgb]{.7,.9,.9}&57.35\quad&90.76\quad&92.67\quad&88.39\quad&00.00\cellcolor[rgb]{.7,.9,.9}&67.95\quad\\
 	    %&\checkmark&\quad&\quad&\checkmark&90.48\quad&78.12\quad&75.72\quad&00.00\cellcolor[rgb]{.7,.9,.9}&61.08\quad&85.19\quad&90.90\quad&88.05\quad&00.00\cellcolor[rgb]{.7,.9,.9}&66.03\quad\\
 	    &(d)\quad&\checkmark&\checkmark&\checkmark&\quad&92.16\quad&82.05\quad&82.51\quad&78.30\cellcolor[rgb]{.7,.9,.9}&83.76\quad&85.25\quad&89.68\quad&86.89\quad&62.52\cellcolor[rgb]{.7,.9,.9}&81.09\quad\\
 	    &(e)\quad&\checkmark&\checkmark&\quad&\checkmark&90.76\quad&82.42\quad&81.00\quad&79.62\cellcolor[rgb]{.7,.9,.9}&83.45\quad&92.29\quad&90.04\quad&88.61\quad&63.33\cellcolor[rgb]{.7,.9,.9}&83.57\quad\\
 	    &(f)\quad&\checkmark&\quad&\checkmark&\checkmark&89.95\quad&81.55\quad&80.89\quad&00.00\cellcolor[rgb]{.7,.9,.9}&63.10\quad&89.98\quad&93.22\quad&89.59\quad&00.00\cellcolor[rgb]{.7,.9,.9}&68.20\quad\\
 	    &(g)\quad&\checkmark&\checkmark&\checkmark&\checkmark&93.37\quad&84.04\quad&85.06\quad&84.63\cellcolor[rgb]{.7,.9,.9}&86.78\quad&93.56\quad&92.49\quad&88.51\quad&69.82\cellcolor[rgb]{.7,.9,.9}&86.10\quad\\
 	    \cline{1-16}
 	    \toprule[1pt]
 	    \bottomrule[1pt]
 	    \multirow{7}{*}{\shortstack{\bf{Collective Results}\\\bf{of Unseen Classes}}}
 	    &(a)\quad&\multicolumn{4}{c|}{Baseline}&38.06\quad&00.10\quad&00.00\quad&00.36\quad&9.63\quad&2.03\quad&1.67\quad&00.00\quad&00.08\quad&00.95\quad\\
 	    \cline{3-6}
 	    &(b)\quad&\quad&\quad&\quad&\quad&1.18\quad\quad&00.00\quad\quad&00.00\quad&00.00\quad&00.30\quad&00.48\quad&00.00\quad&00.00\quad&00.00\quad&00.12\quad\\
 	    &(c)\quad&\checkmark&\quad&\quad&\quad&00.00\quad\quad&00.00\quad\quad&00.00\quad\quad&00.00\quad&00.00\quad&00.00\quad&00.00\quad&00.00\quad&00.00\quad&00.00\quad\\
 	    %&\checkmark &\checkmark&\quad&\quad&77.38\quad&62.34\quad&63.24\quad&67.39\quad &67.59\quad&63.31\quad&78.41\quad&75.99\quad&53.88\quad &67.90\quad\\
 	  %&\checkmark&\quad&\checkmark&\quad&00.00\quad\quad&00.00\quad\quad&00.00\quad\quad&00.00\quad&00.00\quad&00.00\quad&00.00\quad&00.00\quad&00.00\quad&00.00\quad\\
 	  %&\checkmark&\quad&\quad&\checkmark&00.00\quad\quad&00.00\quad\quad&00.00\quad\quad&00.00\quad&00.00\quad&00.00\quad&00.00\quad&00.00\quad&00.00\quad&00.00\quad\\
 	    &(d)\quad&\checkmark&\checkmark&\checkmark&\quad&85.55\quad&76.15\quad&76.78\quad&78.30\quad&79.20\quad&73.09\quad&81.42\quad&82.56\quad&62.52\quad&74.90\quad\\
 	    &(e)\quad&\checkmark&\checkmark&\quad&\checkmark&81.26\quad\quad&75.91\quad\quad&72.24\quad\quad&79.62\quad&77.26\quad&75.49\quad&82.78\quad&82.99\quad&63.33\quad&76.15\quad\\
 	    &(f)\quad&\checkmark&\quad&\checkmark&\checkmark&00.00\quad\quad&00.00\quad\quad&00.00\quad\quad&00.00\quad&00.00\quad&00.00\quad&00.00\quad&00.00\quad&00.00\quad&00.00\quad\\
 	    &(g)\quad&\checkmark&\checkmark&\checkmark&\checkmark&90.61\quad&82.09\quad&82.33\quad&84.63\quad&84.92\quad&84.66\quad&87.55\quad&86.92\quad&69.82\quad&82.24\quad\\
     	\toprule[1pt]
     	\bottomrule[1pt]
        \multicolumn{6}{C{15.0cm}|}{\multirow{1}{*}{\shortstack{\bf{Lower bound}}}}&71.61\quad&54.37\quad&48.29\quad&61.98\quad&59.06\quad&52.14\quad&66.15\quad&67.81\quad&43.18\quad&57.32\\
     	\toprule[1pt]
     	\bottomrule[1pt]
	\end{tabular}
	}
\end{table*}

\begin{figure*}[!htb]
	\centering
	\includegraphics[width=.955\textwidth]{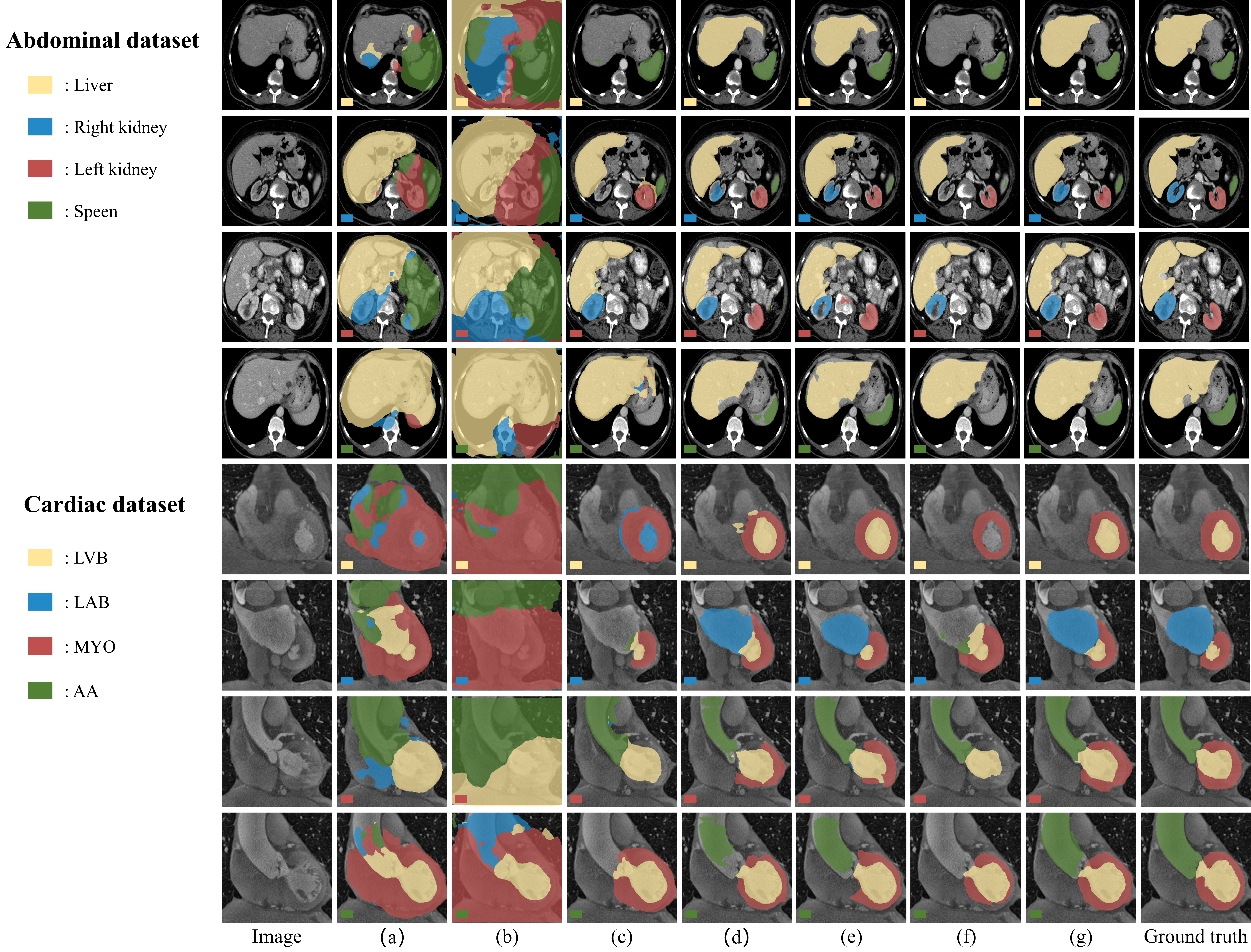}
	\caption{\small
	Visual examples of the ablation study results on the abdominal dataset (top) and the cardiac dataset (bottom);
	the ablation settings (a)--(g) are the same as in Table \ref{table:module_ablations}.
	The color block on the bottom-left corner of the image means the corresponding structure is specified as the unseen classes.
	}%{\color{red}Our proposed modules clearly improve segmentation of the unseen classes for the cardiac and the abdominal structures segmentation.}
	\label{fig:ablation_cardiac}
\end{figure*}

\subsection{Implementation Details}
All experiments are conducted
%The proposed zero-shot segmentation and the prior models are both implemented
with PyTorch\footnote[1]{https://pytorch.org/} on an NVIDIA Tesla P40 GPU of 24 GB memory. We adopt the modified PSP~\cite{zhao2017pyramid} with JPU architecture~\cite{wu2019rethinking} as our backbone for both the prior and zero-shot models, for its powerful feature extraction capability in capturing the contextual information. Meanwhile, PatchGAN~\cite{isola2017image} is selected as the discriminator $\bm{D}$.
The feature fusion function $\bm{M}$ in Eq. (\ref{eq:attention}) is implemented as three convolutional layers, each followed by batch normalization and ReLU.
%using successively three combination modules which consist of convolution layer, batch normalization layer and RELU layer.
The initial learning rate of both models is set to $2.5\times 10^{-4}$, and the SGD optimizer with a weight decay of $5\times 10^{-4}$ is adopted for optimization.
As to $\bm{D}$, the initial learning rate is set to $1\times 10^{-4}$ and the Adam optimizer is chosen for adversarial training with $\beta$'s set to 0.9 and 0.99.
Hyper-parameters in Eq.~(\ref{eq:obj}) (\ie, $\omega_{0}$, $\omega_{1}$, $\omega_{2}$ and $\omega_{3}$) are empirically set to 0.5, 1, 0.01 and 1, respectively.
All volumes are sliced into 2D images for training and testing. The batch size is set to 8 in training. To quantitatively evaluate the performance of the models, Dice score and average symmetric surface distance (ASSD) are employed as the evaluation metrics.
%{\color{red}The importance factors of the probability maps of the unseen class and background class will be set to 0.2 and -0.2 respectively during the inference stage so as to improve the influence of the unseen class.}

\subsection{Experimental Results}

\subsubsection{Analysis of proposed modules}
As presented in Table~\ref{table:module_ablations}, we conduct comprehensive ablation studies on the two datasets.
To verify whether the effectiveness of the proposed modules is related to a specific unseen class, we perform four groups of experiments where each anatomical structure is specified as the only unseen class in turn.
%types of module ablations by specifying each anatomical structure as the only unseen class.
%{\color{red}The baseline in the table denotes that the zero-shot model is finetuned by the well-trained prior model, which can be seen that the relation inheritance procedure is performed, and then is trained with $\mathcal{D}_s$ in a fully supervised manner.}
The baseline in the table denotes that the zero-shot model is finetuned on $\mathcal{D}_s$ in a fully supervised manner, from the well-trained prior model (note this baseline is different from the lower-bound model).
For other experiments, the zero-shot model is jointly trained with the checked components/modules from scratch using $\mathcal{D}_s$.
To facilitate straightforward comparisons across the lower bound, baseline, and different ablation settings, we also collect the results of different unseen classes and present them together (``Collective Results of Unseen Classes'').

% All ablation studies' performance is increased significantly,
From Table~\ref{table:module_ablations}, we make the following observations.
First of all, while the lower-bound performance is far from satisfactory (mean Dice scores of 59.06\% and 57.32\% on the two datasets, respectively), fine-tuning with the annotations of the seen classes (the baseline in row (a)) leads to not only catastrophic forgetting of the unseen classes, but also dramatic decreases in segmentation performance of the seen classes.
We speculate that forcing the baseline model to segment the unseen classes without any annotation severely interferes learning of the seen classes during the fine-tuning process (note this is different from training a model to only segment the seen classes while totally ignoring the unseen classes, which is expected to yield reasonable performance on the seen classes).
These results indicate the challenge of the problem targeted in this work.

Second, without $\mathcal{L}_{Bg}$, the proposed framework cannot function normally (row (b) vs. (c)), clearly demonstrating the importance of the background awareness.
Therefore, we include $\mathcal{L}_{Bg}$ for all subsequent experiments.
Third, without the RPA module, the Dice scores for the unseen classes drop to 0.00\% (rows (c) and (f)), indicating its key role in anti-forgetting.
Besides, removing the RPA module from the proposed framework (row (f) vs. (g)) results in slight to substantial decreases in the segmentation performance of the seen classes in most cases, suggesting that the RPA module also helps the inheritance of seen classes.
This is expected, because the adversarial training aligns the distributions as a whole, without differentiating seen or unseen classes.
Fourth, removing either the IA (row (d)) or the CMA (row (e)) module leads to modest decreases in the segmentation performance of both the seen and unseen classes, but not the catastrophic forgetting problem.
This verifies that: 1) the IA module enables the model to exploit the relation prototypes thoroughly via the attentive inheritance guidance $G$ to extract more representative features for better inheritance, and 2) the CMA module can effectively exploit the rich information of the seen classes contained in the relation prototypes to align features output by the prior and zero-shot models.
Last but not least, the proposed framework achieves substantial improvements upon the lower-bound model when using different classes as the unseen class, suggesting that the effectiveness of the proposed modules is independent of any specific unseen class.
Representative results of the module ablation studies are shown in Fig.~\ref{fig:ablation_cardiac}.

Based on the mean results averaged across the collective results of the unseen classes, we can make straightforward comparisons with the lower bound.
As we can see, our framework improves upon the lower-bound model remarkably by 25.86\% and 24.92\% in absolute Dice scores on the two datasets, respectively.
These results demonstrate the effectiveness of combining the various novel modules in the proposed framework in cross-modal zero-shot medical image segmentation.

%Lastly, due to the catastrophic forgetting problem, the relation prototypes inevitably disturb the well-learned features from seen classes, thus slight drops in performance for seen classes are observed compared to the baseline model. However, when adding the proposed modules incrementally, such performance drops are mitigated.

\begin{table*}[htb]
    \caption{Ablation studies on unseen classes on the abdominal dataset. Dice: Dice coefficient in percentage (\%); ASSD: average symmetric surface distance in millimeters.
         \label{table:abdominal_unseen_ablation}}
    	\centering
    	\renewcommand{\arraystretch}{0.9}
    	\scalebox{0.40}{
    	\LARGE
    	\begin{tabular}{C{7.0cm}|C{2.0cm}|C{2.0cm}|C{2.0cm}|C{2.0cm}|C{2.0cm}|C{2.0cm}|C{2.0cm}|C{2.0cm}|C{2.0cm}|C{2.0cm}|C{2.0cm}|C{2.0cm}|C{2.0cm}|C{2.0cm}}
    		\toprule[1pt]
    		\toprule[1pt]
            \multirow{2}{*}{\shortstack{\bf{Number of}\\\\\bf{Unseen classes}}}
            &\multicolumn{4}{C{7.0cm}|}{\multirow{1}{*}{\shortstack{\bf{Unseen classes}}}}
            &\multicolumn{2}{C{4.0cm}|}{\multirow{1}{*}{\shortstack{\bf{Liver}}}}
     		&\multicolumn{2}{C{4.0cm}|}{\multirow{1}{*}{\shortstack{\bf{R-Kid}}}}
     		&\multicolumn{2}{C{4.0cm}|}{\multirow{1}{*}{\shortstack{\bf{L-Kid}}}}
     		&\multicolumn{2}{C{4.0cm}|}{\multirow{1}{*}{\shortstack{\bf{Speen}}}}
     		&\multicolumn{2}{C{4.0cm}}{\multirow{1}{*}{\shortstack{\bf{Mean}}}}\\
     		\cline{2-15}	
     		&\multicolumn{1}{C{2.0cm}|}{\multirow{1}{*}{\shortstack{\bf{Liver}}}}
     		&\multicolumn{1}{C{2.0cm}|}{\multirow{1}{*}{\shortstack{\bf{R-Kid}}}}
     		&\multicolumn{1}{C{2.0cm}|}{\multirow{1}{*}{\shortstack{\bf{L-Kid}}}}
     		&\multicolumn{1}{C{2.0cm}|}{\multirow{1}{*}{\shortstack{\bf{Speen}}}}&\textbf{Dice}&\textbf{ASSD}&\textbf{Dice}&\textbf{ASSD}&\textbf{Dice}&\textbf{ASSD}&\textbf{Dice}&\textbf{ASSD}&\textbf{Dice}&\textbf{ASSD}\\
     		\toprule[1pt]
     	    \bottomrule[1pt]	
            \multirow{4}{*}{\shortstack{\\\\\\\\\bf{\#1}}}&\checkmark &\quad&\quad&\quad&90.61\quad&1.38\quad&84.15\quad&1.69\quad&86.08\quad&1.37\quad&88.95\quad&1.06\quad&87.45\quad&1.38\quad\\
            &\quad &\checkmark&\quad&\quad&92.42\quad&1.31\quad&82.09\quad&2.10\quad&83.45\quad&1.54\quad&87.53\quad&1.12\quad&86.38\quad&1.52\quad\\
            &\quad &\quad&\checkmark&\quad&93.50\quad&1.12\quad&84.52\quad&1.70\quad&82.33\quad&1.66\quad&89.02\quad&0.99\quad&87.34\quad&1.37\quad\\
            &\quad &\quad&\quad&\checkmark&93.37\quad&1.09\quad&84.04\quad&2.18\quad&85.06\quad&1.61\quad&84.63\quad&1.07\quad&86.78\quad&1.33\quad\\
            \toprule[1pt]
     	    \bottomrule[1pt]
            \multirow{6}{*}{\shortstack{\\\\\\\\\bf{\#2}}}&\checkmark &\checkmark&\quad&\quad&86.24\quad&2.26\quad&77.49\quad&1.87\quad&85.28\quad&1.66\quad&88.89\quad&1.08\quad&84.47\quad&1.72\quad\\
            &\checkmark &\quad&\checkmark&\quad&87.21\quad&2.37\quad&84.78\quad&1.88\quad&77.25\quad&2.56\quad&86.49\quad&1.26\quad&83.93\quad&2.02\quad\\
            &\checkmark &\quad&\quad&\checkmark&86.20\quad&2.11\quad&84.52\quad&1.49\quad&84.57\quad&1.52\quad&82.96\quad&1.56\quad&84.56\quad&1.67\quad\\
            &\quad &\checkmark&\checkmark&\quad&93.15\quad&1.09\quad&77.79\quad&2.18\quad&78.93\quad&1.61\quad&88.19\quad&1.07\quad&84.52\quad&1.49\quad\\
            &\quad &\checkmark&\quad&\checkmark&94.03\quad&1.03\quad&77.51\quad&2.23\quad&84.19\quad&1.46\quad&81.97\quad&1.55\quad&84.43\quad&1.57\quad\\
            &\quad &\quad&\checkmark&\checkmark&92.99\quad&1.21\quad&84.33\quad&1.45\quad&79.31\quad&1.48\quad&83.11\quad&1.39\quad&84.93\quad&1.38\quad\\
            \toprule[1pt]
     	    \bottomrule[1pt]
            \multirow{4}{*}{\shortstack{\\\\\\\\\\\bf{\#3}}}&\checkmark &\checkmark&\checkmark&\quad&87.70\quad&1.92\quad&77.22\quad&1.90\quad&74.63\quad&2.20\quad&88.91\quad&1.02\quad&82.11\quad&1.76\quad\\
            &\checkmark &\checkmark&\quad&\checkmark&84.91\quad&2.44\quad&77.38\quad&2.21\quad&85.51\quad&1.50\quad&81.18\quad&1.77\quad&82.25\quad&1.98\quad\\
             &\checkmark &\quad&\checkmark&\checkmark&87.10\quad&1.94\quad&83.83\quad&1.71\quad&72.73\quad&2.45\quad&80.65\quad&2.09\quad&81.08\quad&2.05\quad\\
            &\quad &\checkmark&\checkmark&\checkmark&93.39\quad&1.25\quad&75.86\quad&2.03\quad&75.10\quad&2.25\quad&79.29\quad&1.83\quad&80.91\quad&1.84\quad\\
    		\bottomrule[1pt]
    		\bottomrule[1pt]
    	\multirow{2}{*}{\shortstack{\bf{Benchmark}}}&\multicolumn{4}{C{7.0cm}|}{\multirow{1}{*}{\shortstack{\bf{Lower bound }}}}&71.61\quad&4.08\quad&54.37\quad&2.95\quad&48.29\quad&3.36\quad&61.98\quad&2.73\quad&59.06\quad&3.28\\
    	
     	&\multicolumn{4}{C{7.0cm}|}{\multirow{1}{*}{\shortstack{\bf{Oracle}}}}&94.91\quad&0.83\quad&87.76\quad&0.99\quad&88.00\quad&1.05\quad&89.83\quad&1.05\quad&90.13\quad&0.95\\
    		\bottomrule[1pt]
    		\bottomrule[1pt]
    	\end{tabular}}
    \end{table*}

 \begin{table*}[htb]
    \caption{Ablation studies on unseen classes on the cardiac dataset. Dice: Dice coefficient in percentage (\%); ASSD: average symmetric surface distance in millimeters.
         \label{table:cardiac_unseen_ablation}}
    	\centering
    	\renewcommand{\arraystretch}{0.9}
    	\scalebox{0.40}{
    	\LARGE
    	\begin{tabular}{C{7.0cm}|C{2.0cm}|C{2.0cm}|C{2.0cm}|C{2.0cm}|C{2.0cm}|C{2.0cm}|C{2.0cm}|C{2.0cm}|C{2.0cm}|C{2.0cm}|C{2.0cm}|C{2.0cm}|C{2.0cm}|C{2.0cm}}
    		\toprule[1pt]
    		\toprule[1pt]
            \multirow{2}{*}{\shortstack{\bf{Number of}\\\\\bf{Unseen classes}}}
            &\multicolumn{4}{C{7.0cm}|}{\multirow{1}{*}{\shortstack{\bf{Unseen classes}}}}
            &\multicolumn{2}{C{4.0cm}|}{\multirow{1}{*}{\shortstack{\bf{AA}}}}
     		&\multicolumn{2}{C{4.0cm}|}{\multirow{1}{*}{\shortstack{\bf{LAB}}}}
     		&\multicolumn{2}{C{4.0cm}|}{\multirow{1}{*}{\shortstack{\bf{LVB}}}}
     		&\multicolumn{2}{C{4.0cm}|}{\multirow{1}{*}{\shortstack{\bf{MYO}}}}
     		&\multicolumn{2}{C{4.0cm}}{\multirow{1}{*}{\shortstack{\bf{Mean}}}}\\
     		\cline{2-15}	
     		&\multicolumn{1}{C{2.0cm}|}{\multirow{1}{*}{\shortstack{\bf{AA}}}}
     		&\multicolumn{1}{C{2.0cm}|}{\multirow{1}{*}{\shortstack{\bf{LAB}}}}
     		&\multicolumn{1}{C{2.0cm}|}{\multirow{1}{*}{\shortstack{\bf{LVB}}}}
     		&\multicolumn{1}{C{2.0cm}|}{\multirow{1}{*}{\shortstack{\bf{MYO}}}}&\textbf{Dice}&\textbf{ASSD}&\textbf{Dice}&\textbf{ASSD}&\textbf{Dice}&\textbf{ASSD}&\textbf{Dice}&\textbf{ASSD}&\textbf{Dice}&\textbf{ASSD}\\
     		\toprule[1pt]
     	    \bottomrule[1pt]	
            \multirow{4}{*}{\shortstack{\\\\\\\\\bf{\#1}}}&\checkmark &\quad&\quad&\quad&84.66\quad&4.06\quad&91.63\quad&2.48\quad&89.20\quad&2.73\quad&80.92\quad&2.60\quad&86.60\quad&2.97\quad\\
            &\quad &\checkmark&\quad&\quad&93.06\quad&2.08\quad&87.55\quad&3.73\quad&88.64\quad&2.90\quad&79.34\quad&2.71\quad&87.15\quad&2.85\quad\\
            &\quad &\quad&\checkmark&\quad&92.74\quad&2.22\quad&92.66\quad&2.23\quad&86.92\quad&3.73\quad&79.52\quad&2.63\quad&87.96\quad&2.70\quad\\
            &\quad &\quad&\quad&\checkmark&93.56\quad&1.70\quad&92.49\quad&2.25\quad&88.51\quad&2.71\quad&69.82\quad&4.18\quad&86.10\quad&2.71\quad\\
            \toprule[1pt]
     	    \bottomrule[1pt]
            \multirow{6}{*}{\shortstack{\\\\\\\\\bf{\#2}}}&\checkmark &\checkmark&\quad&\quad&78.55\quad&5.74\quad&86.64\quad&4.42\quad&87.77\quad&3.32\quad&80.10\quad&2.60\quad&83.27\quad&4.02\quad\\
            &\checkmark &\quad&\checkmark&\quad&81.60\quad&4.36\quad&92.23\quad&2.32\quad&84.78\quad&4.53\quad&76.85\quad&2.83\quad&83.87\quad&3.51\quad\\
            &\checkmark &\quad&\quad&\checkmark&81.20\quad&4.35\quad&91.10\quad&2.28\quad&89.50\quad&2.85\quad&66.20\quad&4.40\quad&82.00\quad&3.47\quad\\
            &\quad &\checkmark&\checkmark&\quad&92.72\quad&1.85\quad&85.98\quad&4.04\quad&85.83\quad&3.95\quad&79.41\quad&2.82\quad&85.98\quad&3.17\quad\\
            &\quad &\checkmark&\quad&\checkmark&92.73\quad&1.75\quad&86.80\quad&3.86\quad&87.27\quad&2.98\quad&65.80\quad&4.10\quad&83.15\quad&3.17\quad\\
            &\quad &\quad&\checkmark&\checkmark&93.69\quad&1.86\quad&93.14\quad&2.11\quad&83.20\quad&4.12\quad&59.95\quad&4.94\quad&82.50\quad&3.26\quad\\
            \toprule[1pt]
     	    \bottomrule[1pt]
            \multirow{4}{*}{\shortstack{\\\\\\\\\\\bf{\#3}}}&\checkmark &\checkmark&\checkmark&\quad&78.13\quad&5.79\quad&81.11\quad&3.83\quad&79.30\quad&4.74\quad&76.46\quad&3.97\quad&78.74\quad&4.58\quad\\
            &\checkmark &\checkmark&\quad&\checkmark&78.81\quad&5.10\quad&80.58\quad&4.60\quad&87.09\quad&2.97\quad&63.95\quad&4.27\quad&77.62\quad&4.24\quad\\
             &\checkmark &\quad&\checkmark&\checkmark&77.27\quad&5.63\quad&91.78\quad&2.81\quad&76.09\quad&5.14\quad&59.48\quad&4.99\quad&76.16\quad&4.64\quad\\
            &\quad &\checkmark&\checkmark&\checkmark&90.40\quad&1.96\quad&81.60\quad&3.80\quad&79.40\quad&4.36\quad&60.31\quad&5.29\quad&77.93\quad&3.86\quad\\
    		\bottomrule[1pt]
    		\bottomrule[1pt]
    		 \multirow{2}{*}{\shortstack{\bf{Benchmark}}}&\multicolumn{4}{C{7.0cm}|}{\multirow{1}{*}{\shortstack{\bf{Lower bound }}}}&52.14\quad&9.50\quad&66.15\quad&8.52\quad&67.81\quad&6.56\quad&43.18\quad&6.73\quad&57.32\quad&7.83\\
    	     &\multicolumn{4}{C{7.0cm}|}{\multirow{1}{*}{\shortstack{\bf{Oracle}}}}&95.04\quad&1.21\quad&93.56\quad&1.86\quad&89.97\quad&2.26\quad&83.43\quad&2.25\quad&90.05\quad&1.89\\
    		\bottomrule[1pt]
    		\bottomrule[1pt]
    	\end{tabular}}
    \end{table*}

\begin{table*}[htb]
    \caption{Comparisons of the SOTA UDA algorithms and the proposed zero-shot segmentation framework trained with different numbers of unseen classes.
    Note that for zero-shot methods (ours and the ZS3Net) each class would be chosen as the unseen class to compare with the UDA methods. For our framework with more than one unseen class, the results are reported as the average performance of those models that specify the same class as unseen class. Dice score is used as the evaluation metric and reported in the percentage (\%). \label{table:quanti_metric}}
    	\centering
    	\renewcommand{\arraystretch}{1.0}
    	\scalebox{0.46}{
    	\LARGE
    	\begin{tabular}{C{4.5cm}|C{4.5cm}|C{1.7cm}|C{1.7cm}|C{1.7cm}|C{1.7cm}|C{1.7cm}|C{1.7cm}|C{1.7cm}|C{1.7cm}|C{1.7cm}|C{1.7cm}}
    		\toprule[1pt]
    		\toprule[1pt]
            \multicolumn{1}{C{4.5cm}|}{\multirow{2}{*}{\bf{Approach}}}&\multicolumn{1}{C{4.5cm}|}{\multirow{2}{*}{\bf{Method}}}&\multicolumn{5}{C{8.5cm}|}{\bf{Abdominal}}& \multicolumn{5}{C{8.5cm}}{\bf{Cardiac}}\\
            \cline{3-12}	
     		&&\multicolumn{1}{C{1.7cm}|}{\multirow{1}{*}{\shortstack{\bf{Liver}}}}
        	&\multicolumn{1}{C{1.7cm}|}{\multirow{1}{*}{\shortstack{\bf{R-Kid}}}}
     		&\multicolumn{1}{C{1.7cm}|}{\multirow{1}{*}{\shortstack{\bf{L-Kid}}}}
     		&\multicolumn{1}{C{1.7cm}|}{\multirow{1}{*}{\shortstack{\bf{Speen}}}}
     		&\multicolumn{1}{C{1.7cm}|}{\multirow{1}{*}{\shortstack{\bf{Mean}}}}
     		&\multicolumn{1}{C{1.7cm}|}{\multirow{1}{*}{\shortstack{\bf{AA}}}}
        	&\multicolumn{1}{C{1.7cm}|}{\multirow{1}{*}{\shortstack{\bf{LAB}}}}
     		&\multicolumn{1}{C{1.7cm}|}{\multirow{1}{*}{\shortstack{\bf{LVB}}}}
     		&\multicolumn{1}{C{1.7cm}|}{\multirow{1}{*}{\shortstack{\bf{MYO}}}}
     		&\multicolumn{1}{C{1.7cm}}{\multirow{1}{*}{\shortstack{\bf{Mean}}}}\\
     		\cline{3-12}	
     		\toprule[1pt]
    		\toprule[1pt]
          \multicolumn{1}{C{4.5cm}|}{\multirow{5}{*}{\bf{UDA}}}&\textbf{AdaptSegNet~\cite{tsai2018learning}}&82.65\quad&51.52\quad&29.21\quad&50.34\quad&53.43\quad&65.35\quad&80.63\quad&81.43\quad&69.34\quad&74.18\quad\\
          &\textbf{CLAN~\cite{luo2019taking}}&74.64\quad&53.93\quad&41.93\quad&53.22\quad&55.93\quad&63.81\quad&79.93\quad&84.40\quad&66.76\quad&73.73\quad\\
          &\textbf{BDL~\cite{li2019bidirectional}}&74.19\quad&52.36\quad&21.55\quad&45.63\quad&48.43\quad&67.10\quad&80.55\quad&82.68\quad&62.06\quad&73.10\quad\\
          &\textbf{DISE~\cite{chang2019all}}&88.88\quad&80.51\quad&79.12\quad&78.83\quad&81.84\quad&71.75\quad&82.20\quad&83.69\quad&60.75\quad&74.60\quad\\
          &\textbf{SIFA~\cite{chen2020unsupervised}}&88.00\quad&\textbf{83.30}\quad&80.90\quad&82.60\quad&83.70\quad&81.30\quad&79.50\quad&73.80\quad&61.60\quad&74.05\quad\\
          &\textbf{UADA~\cite{bian2020uncertainty}}&84.37\quad&82.62\quad&\textbf{84.48}\quad&84.54\quad&84.00\quad&84.15\quad&\textbf{88.30}\quad&84.32\quad&\textbf{71.42}\quad&82.05\quad\\
          	\bottomrule[1pt]
    		\bottomrule[1pt]
          \multicolumn{1}{C{4.0cm}|}{\multirow{3}{*}{\bf{Zero-Shot}}}&\textbf{Ours \#1}&\textbf{90.61}\quad&82.09\quad&82.33\quad&\textbf{84.63}\quad&\textbf{84.92}\quad&\textbf{84.66}\quad&87.55\quad&\textbf{86.92}\quad&69.82\quad&\textbf{82.24}\quad\\

          &\textbf{Ours \#2}&86.55\quad&77.60\quad&78.50\quad&82.68\quad&81.33\quad&80.45\quad&86.47\quad&84.60\quad&63.98\quad&78.88\quad\\

          &\textbf{Ours \#3}&86.57\quad&76.82\quad&74.15\quad&80.37\quad&79.48\quad&78.07\quad&81.10\quad&78.26\quad&61.25\quad&74.67\quad\\

          &\textbf{{\color{blue}ZS3Net \#1}~\cite{bucher2019zero}}&36.16\quad&34.85\quad&19.10\quad&30.42\quad&30.13\quad&29.17\quad&13.03\quad&32.08\quad&15.24\quad&22.38\quad\\
    		\bottomrule[1pt]
    		\bottomrule[1pt]
    	\multirow{2}{*}{\shortstack{\bf{Benchmark}}}&\multicolumn{1}{C{4.5cm}|}{\multirow{1}{*}{\shortstack{\bf{Lower bound }}}}&71.61\quad&54.37\quad&48.29\quad&61.98\quad&59.06\quad&52.14\quad&66.15\quad&67.81\quad&43.18\quad&57.32\quad\\
    	
     	&\multicolumn{1}{C{4.5cm}|}{\multirow{1}{*}{\shortstack{\bf{Oracle }}}}&94.91\quad&87.76\quad&88.00\quad&89.83\quad&90.13\quad&95.04\quad&93.56\quad&89.97\quad&83.43\quad&90.05\quad\\
    		\bottomrule[1pt]
    		\bottomrule[1pt]
    	\end{tabular}}
    \end{table*}

\subsubsection{Impact of varying number of unseen classes}
%In the previous experiment, the efficacy of our modules have been proven. However, the generalization ability of our framework is still unknown.
To investigate the generalization capability of the proposed framework, we list all possible combinations of unseen classes on the two datasets and evaluate the performance.
The results are presented in Table~\ref{table:abdominal_unseen_ablation} and Table~\ref{table:cardiac_unseen_ablation}.
It is noticed that the overall performance decreases as the number of unseen classes increases.
This is intuitive that the more unseen classes we consider, the less information is provided by the seen classes to define the relation prototypes in Modality B,
%the more chaos information the relation prototype induce
which makes the inheritance more challenging.
% For this reason, when considering single unseen class in the experiment, those results rank the best in both Table~\ref{table:abdominal_unseen_ablation} and Table~\ref{table:cardiac_unseen_ablation}.
Furthermore, the performances of the proposed framework with one, two, and three unseen classes surpass that of the lower-bound model substantially.
In fact, when a single class is specified as unseen, the performance is close to that of Oracle.
Lastly, we observe only slight drops in performance along with the increase in the number of unseen classes.
These results suggest that the proposed framework can well generalize to an arbitrary number of unseen classes.

\subsubsection{Comparison with SOTA UDA and ZSL approaches}
To the best of our knowledge, this work is the first zero-shot segmentation framework for medical images across modalities.
As mentioned in Section~\ref{section:related work_UDA}, the most relevant approaches are the UDA techniques, since both UDA and the proposed framework utilize data from another imaging modality for training.
Though we list the performance of UDA methods in Table \ref{table:quanti_metric}, we should be cautious to compare it directly with that of the proposed zero-shot method. The differences lie within several aspects. Firstly, we adopt the supervised strategy for the training of seen classes in our framework, while the UDA methods are trained without supervision in the target domain. Secondly, the training data of the prior model is not accessible to the proposed zero-shot model. In contrast with the UDA methods which utilize data of two modalities simultaneously for adversarial learning, the only usable resource in our method is the well-trained prior model.
Thirdly, the target of the zero-shot segmentation is to maintain the performance of seen classes given the annotations for supervision and to improve the performance of the unseen classes with the proposed relation prototype inheritance, whereas the UDA methods aim to improve the performance of the target modality as a whole without differentiating seen and unseen classes.

{\color{blue}Additionally, we also include the ZS3Net \cite{bucher2019zero}, a zero-shot semantic segmentation architecture combining a deep visual segmentation model with an approach to generating visual representations from semantic word embeddings (Word2Vec \cite{mikolov2013distributed}), for comparison.}
To adapt ZS3Net for image-based semantic embeddings, we extract in Modality A the {\color{blue}class-wise} prototypes for all classes using the well-trained prior model, and substitute them for the word embeddings in \cite{bucher2019zero}.
Specifically, the prototype for a specific class is computed as the pixel-level mean feature vector averaged over all pixels belonging to that class in $\mathcal{D}_p$, according to the annotations.
To this end, the prototypes are used as the auxiliary prior knowledge for the ZSL, just like the word embeddings.
The ZS3Net is evaluated with one unseen class.

\begin{table*}[htb]
    \caption{Comparisons of the SOTA UDA algorithms and the proposed zero-shot segmentation framework on CT to MRI adaptation.
    Note that for zero-shot methods (ours and the ZS3Net) each class was chosen as the unseen class to compare with the UDA methods. Dice score is used as the evaluation metric and reported in the percentage (\%). \label{table:quanti_metric_reverse}}
    	\centering
    	\renewcommand{\arraystretch}{1.0}
    	\scalebox{0.46}{
    	\LARGE
    	\begin{tabular}{C{4.5cm}|C{4.5cm}|C{1.7cm}|C{1.7cm}|C{1.7cm}|C{1.7cm}|C{1.7cm}|C{1.7cm}|C{1.7cm}|C{1.7cm}|C{1.7cm}|C{1.7cm}}
    		\toprule[1pt]
    		\toprule[1pt]
            \multicolumn{1}{C{4.5cm}|}{\multirow{2}{*}{\bf{Approach}}}&\multicolumn{1}{C{4.5cm}|}{\multirow{2}{*}{\bf{Method}}}&\multicolumn{5}{C{8.5cm}|}{\bf{Abdominal}}& \multicolumn{5}{C{8.5cm}}{\bf{Cardiac}}\\
            \cline{3-12}	
     		&&\multicolumn{1}{C{1.7cm}|}{\multirow{1}{*}{\shortstack{\bf{Liver}}}}
        	&\multicolumn{1}{C{1.7cm}|}{\multirow{1}{*}{\shortstack{\bf{R-Kid}}}}
     		&\multicolumn{1}{C{1.7cm}|}{\multirow{1}{*}{\shortstack{\bf{L-Kid}}}}
     		&\multicolumn{1}{C{1.7cm}|}{\multirow{1}{*}{\shortstack{\bf{Speen}}}}
     		&\multicolumn{1}{C{1.7cm}|}{\multirow{1}{*}{\shortstack{\bf{Mean}}}}
     		&\multicolumn{1}{C{1.7cm}|}{\multirow{1}{*}{\shortstack{\bf{AA}}}}
        	&\multicolumn{1}{C{1.7cm}|}{\multirow{1}{*}{\shortstack{\bf{LAB}}}}
     		&\multicolumn{1}{C{1.7cm}|}{\multirow{1}{*}{\shortstack{\bf{LVB}}}}
     		&\multicolumn{1}{C{1.7cm}|}{\multirow{1}{*}{\shortstack{\bf{MYO}}}}
     		&\multicolumn{1}{C{1.7cm}}{\multirow{1}{*}{\shortstack{\bf{Mean}}}}\\
     		\cline{3-12}	
     		\toprule[1pt]
    		\toprule[1pt]
          \multicolumn{1}{C{4.5cm}|}{\multirow{5}{*}{\bf{UDA}}}&\textbf{AdaptSegNet~\cite{tsai2018learning}}&78.91\quad&78.16\quad&75.92\quad&80.61\quad&78.40\quad&00.13\quad&00.79\quad&64.60\quad&33.91\quad&24.86\quad\\
          &\textbf{CLAN~\cite{luo2019taking}}&77.02\quad&81.66\quad&76.23\quad&82.10\quad&79.25\quad&1.59\quad&15.46\quad&65.71\quad&36.50\quad&29.82\quad\\
          &\textbf{BDL~\cite{li2019bidirectional}}&74.78\quad&76.11\quad&74.25\quad&85.46\quad&77.65\quad&20.61\quad&30.20\quad&47.15\quad&24.97\quad&30.73\quad\\
          &\textbf{DISE~\cite{chang2019all}}&85.26\quad&85.06\quad&85.03\quad&\textbf{90.29}\quad&86.41\quad&39.45\quad&46.13\quad&83.86\quad&55.28\quad&56.18\quad\\
          &\textbf{SIFA~\cite{chen2020unsupervised}}&\textbf{90.00}\quad&\textbf{89.10}\quad&80.20\quad&82.30\quad&85.40\quad&65.30\quad&\textbf{62.30}\quad&78.90\quad&47.30\quad&63.40\quad\\
          &\textbf{UADA~\cite{bian2020uncertainty}}&88.68\quad&85.60\quad&\textbf{85.63}\quad&87.99\quad&\textbf{86.98}\quad&66.11\quad&59.45\quad&83.34\quad&64.39\quad&68.32\quad\\
          	\bottomrule[1pt]
    		\bottomrule[1pt]
          \multicolumn{1}{C{4.0cm}|}{\multirow{3}{*}{\bf{Zero-Shot}}}&\textbf{Ours \#1}&88.39\quad&83.26\quad&82.61\quad&82.18\quad&84.11\quad&\textbf{71.28}\quad&55.02\quad&\textbf{88.20}\quad&\textbf{70.83}\quad&\textbf{71.33}\quad\\

          &\textbf{{\color{blue}ZS3Net \#1}~\cite{bucher2019zero}}&18.47\quad&18.85\quad&32.01\quad&35.89\quad&26.31\quad&6.51\quad&5.76\quad&13.81\quad&8.75\quad&8.71\quad\\
    		\bottomrule[1pt]
    		\bottomrule[1pt]
    	\multirow{2}{*}{\shortstack{\bf{Benchmark}}}&\multicolumn{1}{C{4.5cm}|}{\multirow{1}{*}{\shortstack{\bf{Lower bound }}}}&78.35\quad&65.02\quad&58.14\quad&77.31\quad&69.70\quad&49.00\quad&49.10\quad&84.60\quad&43.03\quad&56.00\quad\\
    	
     	&\multicolumn{1}{C{4.5cm}|}{\multirow{1}{*}{\shortstack{\bf{Oracle }}}}&93.89\quad&93.34\quad&92.30\quad&91.95\quad&92.87\quad&82.28\quad&83.44\quad&91.84\quad&77.62\quad&83.80\quad\\
    		\bottomrule[1pt]
    		\bottomrule[1pt]
    	\end{tabular}}
    \end{table*}

As to our proposed framework, we evaluate three versions which are trained with different numbers of unseen classes.
For fair comparisons with the UDA methods, we only report the performance of unseen classes in Table~\ref{table:quanti_metric}. %, where the result of each class is saved from the experiment when they are specified as the only unseen class.
As we can see, the single unseen class version of our method (Ours \#1) slightly outperforms all SOTA UDA methods, with the absolute improvements of 0.92\% and 0.19\% in cross-organ and cross-tissue mean Dice scores for the abdominal and cardiac datasets, respectively.
Even when the number of unseen classes increases to 2 and 3, the performances of ours (Ours \#2 and \#3) are comparable to most of the competing methods. Moreover, it is observed that the performance of our model is close to that of Oracle, indicating that our method can help alleviate the \textit{domain shift} problem to some extent.
Lastly, ZS3Net is apparently unsuitable for the problem targeted in this work, yielding even inferior performance to the lower bound.

\section{Discussion and Conclusion}
In this paper, we raised a new zero-shot problem for multi-modality segmentation in medical image analysis, the solution to which was aimed to address the public concern when facing both the data privacy and shortage issues at the same time.
Then, we proposed a novel annotation-efficient medical image segmentation framework to address this problem, which can segment unseen target structures by utilizing the prior model and annotations of seen classes.
To the best of our knowledge, we were the first to apply the concept of zero-shot learning (ZSL) to the multimodal imaging scenario.
The proposed method consisted of two stages: the relation prototype training stage and the relation inheritance training stage.
To realize ZSL, we extracted the relation prototypes from the prior model and transferred them to the zero-shot model.
More specifically, we proposed the cross-modality adaptation (CMA) module, relation prototype awareness (RPA) module, and inheritance attention (IA) module to achieve the inheritance of the relation prototypes and boost the zero-shot performance.
Our framework achieved competitive performance compared with existing SOTA unsupervised domain adaptation (UDA) methods on two cross-modality medical image datasets, including the abdominal dataset and the MMWHS cardiac dataset. In addition, we explored the effects of the number of unseen classes and different unseen class selections on our proposed framework.
The experimental results showed that our approach was robust and effective.

One of the major contributions of our proposed framework was that it achieved performances better than the SOTA UDA approaches without using data from the prior modality. In this sense, the proposed framework is a more practical solution than UDA, since it only requires the source domain model as a prior to train the target ZSL model so that the problem of the source data privacy can be avoided.
%However, a limitation is that the prior knowledge is extracted only from the images, whereas the medical reports as the prior knowledge contain rich information, which are expected to further contribute to the ZSL. In the future, more effort will be made to investigate the heterogeneous modality (\eg text and image) zero-shot learning so as to improve the generalization of our model.

In essence, this work applied the concept of ZSL to the problem of partially supervised domain adaptation (DA) for annotation-efficient cross-modal medical image segmentation.
%The investigated problem setting was between unsupervised and fully supervised DA in that not all of the segmentation classes were annotated for the target modality.
Accordingly, we identified two challenges that needed to be addressed for successful application.
First, it is widely accepted that the projection domain shift problem \cite{fu2015transductive,lazaridou2015hubness} is the fundamental problem in ZSL.
To tackle this problem, we proposed the CMA module to calibrate the common projection semantic space of two models, such that the relation prototypes of the seen classes previously learned by the prior model were effectively inherited by the zero-shot model.
Second, as only the seen classes were annotated in the target modality for supervision, the zero-shot model would forget the established prior knowledge about the unseen classes if not taken care of. Borrowing from the continual learning literature~\cite{mccloskey1989catastrophic,ratcliff1990connectionist,kirkpatrick2017overcoming,douillard2021insights}, we called this phenomenon ``catastrophic forgetting''---the tendency of an artificial neural network to forget previously learned information upon learning new information.
%is often referred to as ``catastrophic forgetting''---the tendency of an artificial neural network to completely and abruptly forget previously learned information upon learning new information, and is one of the basic problems in the continual learning literature \textcolor{red}{[survey]}.
Accordingly, we proposed the RPA module which implicitly encouraged the zero-shot model to output features for the unseen classes that were indistinguishable from those output by the prior model via adversarial training.
Our ablation study (Table \ref{table:module_ablations}) demonstrated the effectiveness of the two modules.

{\color{blue}Compared to the ZSL of natural images, we identified two notable differences.
First, while most existing ZSL approaches for natural images relied on word embeddings (such as the WordNet~\cite{miller1995wordnet}) and manual attributes, such linguistic models are currently unavailable for medical images due to the professionalism of medical terminology.
To circumvent this issue, we innovatively proposed to exploit cross-modal image prior as the auxiliary information in zero-shot segmentation of multimodal medical images.
Second, the definition of unseen classes is distinct:
as multimodal medical images of the same body part often capture consistent structures, ``unseen'' here actually meant unannotated classes, instead of classes not seen before.
Furthermore, the intensity mappings between medical images of different modalities are usually nonlinear, which further complicates the problem.
These two intertwining factors made the utilization of cross-modal image prior for ZSL of medical images a challenging problem.
In fact, the straightforward adaption of the ZS3Net~\cite{bucher2019zero}, a representative zero-shot semantic segmentation method for natural images, failed to produce meaningful results in this challenging new task (Table \ref{table:quanti_metric} and Table~\ref{table:quanti_metric_reverse}).
In contrast, our framework yielded competitive performance against several SOTA UDA methods, with the ablation studies verifying the effectiveness of its novel elements, including: $\mathcal{L}_{Bg}$, RPA, CMA, and IA.}

For experiments, this study followed the convention in the UDA literature~\cite{chen2019synergistic,bian2020uncertainty} where the adaptation direction was from MRI to CT.
In clinical practice, however, the access to MRI is more limited than to CT, resulting in fewer scans available for the former than for the latter.
In addition, MRI can be more difficult to process owing to non-standard images across scanners, vendors, institutions, and even patients, whereas CT measures Hounsfield units which are more consistent.
Therefore, it is expected to make a greater impact if the adaptation direction is reversed, \emph{i.e.}, from CT to MRI, and the proposed framework would still work.
For this reason, we conducted extra exploratory experiments on both the abdominal and cardiac datasets, where the prior model was trained on the CT data and the zero-shot model was trained and tested on the MRI data.
The results were shown in Table \ref{table:quanti_metric_reverse}, which demonstrated trends similar to the results of MRI-to-CT adaptation.
%In general, the performance of most of the UDA and ZSL approaches improved upon the lower bound on at least either one of the two datasets, demonstrating the overall benefits of domain adaptation.
Firstly, our framework (with one unseen class) improved substantially upon the lower-bound model on both datasets, with absolute increases of 14.41\% and 15.33\% in cross-structure mean Dice scores, respectively.
Secondly, it achieved comparable results with the most competitive UDA methods on the abdominal dataset, and the best mean Dice score on the cardiac dataset.
Thirdly, our framework was among the only three approaches that improved on the lower-bound model on both datasets.
These results demonstrated the effectiveness, potential, and competency of our framework in the practically more impactful adaptation direction.

We identified a few directions for future work.
First, the prior knowledge utilized in this work for ZSL was extracted from image data of a different modality.
In the future, we plan to study further integration of the prior knowledge in the form of medical reports, when linguistic embedding models for professional medical languages become available.
Second, besides semantic segmentation, object detection---another fundamental task in computer vision---also has important applications in medical image analysis such as lesion detection \cite{xie2021recist}.
Hence, future work should also be devoted to development of zero-shot detection methodologies \cite{bansal2018zero} for cross-modal medical image analysis.

This work had limitations.
One was the implicit assumption underlying our framework, that the same set of structures should be present across patients.
%On one hand, this assumption is expected to hold in most cases considering the rareness of exceptions.
While unusual, patients can miss organs in clinical practice (\emph{e.g.}, after nephrectomy).
It would be meaningful to study the impact of such cases on the proposed framework in the future.

%\IEEEtriggeratref{46}
\bibliographystyle{IEEEtran}
\bibliography{zsv2}

\end{document}